\documentclass[lettersize,journal]{IEEEtran}
\usepackage{amsmath,amsfonts}
\usepackage{algorithmic}
\usepackage{algorithm}
\usepackage{array}
\usepackage[caption=false,font=normalsize,labelfont=sf,textfont=sf]{subfig}
\usepackage{textcomp}
\usepackage{stfloats}
\usepackage{url}
\usepackage{verbatim}
\usepackage{graphicx}
\usepackage{hyperref}
\usepackage{cite}
\usepackage{pdflscape}
\usepackage{multirow}
\usepackage{longtable}
\usepackage{xltabular}
\begin{document}

\title{A Review on Building Blocks of Decentralized Artificial Intelligence}

\author{Vid~Kersic, \textit{University of Maribor},
        and~Muhamed~Turkanovic, \textit{University of Maribor}
\thanks{
Vid Kersic and Muhamed Turkanovic are with the Faculty of Electrical Engineering and Computer Science, University of Maribor, 2000 Maribor Slovenia (e-mail: vid.kersic@um.si, muhamed.turkanovic@um.si). 
}}



\maketitle

\begin{abstract}
Artificial intelligence is transforming our lives, and technological progress and transfer from the academic and theoretical sphere to the real world are accelerating yearly. But during that progress and transition, several open problems and questions need to be addressed for the field to develop ethically, such as digital privacy, ownership, and control. These are some of the reasons why the currently most popular approaches of artificial intelligence, i.e., centralized AI (CEAI), are questionable, with other directions also being widely explored, such as decentralized artificial intelligence (DEAI), to solve some of the most reaching problems. This paper provides a systematic literature review (SLR) of existing work in the field of DEAI, presenting the findings of 71 identified studies. The paper's primary focus is identifying the building blocks of DEAI solutions and networks, tackling the DEAI analysis from a bottom-up approach. In the end, future directions of research and open problems are proposed.
\end{abstract}


\section{Introduction}
\label{sec:introduction}

Since the beginning of the last decade, the development and progress of artificial intelligence (AI) have been advancing faster each year. With many of its subfields and ideas, transitioning from the conceptual and academic sphere towards the applied one (e.g., generative AI), the influence of AI on society is getting more powerful and will be even more in the future. Because of the fast and sudden integration of AI agents and tools into our community, there is a growing concern and debate around AI regulations, availability, governance of training and models, and who should control and own each part of AI development and usage.

Similar to the evolution of the internet and web technologies, AI development followed a similar trajectory with technology and advancements happening in centralized locations with single entities controlling the systems, i.e., big data centers and big tech companies. This is especially present in machine learning (ML) systems, such as large language models (LLM) and ChatGPT, which require huge amounts of data and computing power to achieve state-of-the-art (SOTA) results \cite{vaswani2017attention, jordan2015machine}. Since such solutions are currently mainly provided by big tech players, this type of AI can be referred to as centralized AI (CEAI). While the internet itself is a decentralized and neutral protocol, solutions built on top of it naturally evolved in a centralized way due to the faster and easier adoption, but this resulted in a small number of actors controlling and governing the platforms that are used by millions or billions of people and are integrated into several aspects of our lives \cite{moura2020clouding}. This leads to several problems, such as data of the users being concentrated and owned by platform owners, which in turn leads to data being more vulnerable and easier to steal or be leaked due to being located in a single location. Even though the architecture is distributed, the control of the application and its data still resides in the hands of a single entity. Another problem is the platforms' primary goal, which is to be profitable for the stakeholders, meaning they are often built on the wrong incentives in the first place. Thus, there are many similarities between the internet and AI evolution throughout the decades, with technological advancements strongly relying on owning more data and money.

However, things have started changing in the past few years with the increased development and interest behind decentralized and peer-to-peer (P2P) protocols, such as Bitcoin, Ethereum, IPFS, and Filecoin, which makes it possible to have digital currencies, state machines, and file sharing in a decentralized manner \cite{nakamoto2008bitcoin, buterin2014next, ipfs, filecoin}. One of the leading technologies that have supercharged the development of these networks, though it is not always needed, is blockchain or, more broadly, distributed ledger technologies (DLTs). This new type of networks and systems are often interchangeably referred to as Web3, building on the ethos of (i) \textit{decentralization}, (ii) \textit{self-sovereignty}, (iii) \textit{ownership or controllership of data and processes}, (iv) \textit{no central governance} and/or \textit{no single point of failure} of the systems, as well as (v) \textit{privacy}. Thus, many decentralized solutions started taking some of the market share from their centralized counterparts. 

However, the AI field is still being entirely developed in the CEAI way. While there are several attempts and research in distributing some parts of the AI development, i.e., distributed AI (DAI) and multi-agent systems (MAS), the focus is more on using multiple machines or splitting the systems into numerous parts for computation or efficiency benefits \cite{miiller1990decentralized, chaib1992trends}. But in such solutions, the AI systems are still in the hands of centralized entities, and they are being developed behind walled gardens, which is entirely different from decentralized and P2P networks. In the latter, development happens in the open-source fashion, which has proved to be a powerful tool in the longer horizon in contrast to products developed by centralized entities, which cannot take advantage of such network effects and diverse communities created around technology. Therefore, decentralized solutions are often better and more extensible than centralized solutions. Such solutions can also get faster feedback from the users and the biggest network, which can contribute positively to the result with humans-in-the-loop in the development process. It is also important to emphasize that the DAI field was also called decentralized artificial intelligence (DEAI) \cite{miiller1990decentralized}. Still, the word "decentralized" has been used in the last years for networks and applications with no centralized governance and single point of failure, redefined by the emerging Web3 technologies. Thus, this paper uses DEAI for AI solutions following Web3 paradigms.

As mentioned before, there is no widely adopted decentralized network for AI that would enable interaction, communication, training, inference, and interoperability between AI agents or models. However, some solutions are being researched, developed, or proposed. Using the principles of decentralization, many of the following problems of CEAI would be solved: (i) \textit{verifiability of remotely run models}, (ii) \textit{usability of publicly available AI models}, (iii) \textit{incentivization for AI developers and users}, (iv) \textit{global governance of essential solutions in the digital society}, (v) \textit{no vendor lock-ins}, etc. Decentralization of AI could make the development and progress of AI similar to the internet and other Web3 technologies, where improvements in one part of the technology stack affect and improve products in the layers above (sometimes also below), resulting in positive network effects across the whole ecosystem. The technologies that are being developed without strong influence from a few powerful actors can build more robust and more diversified communities and attract people from different backgrounds (industry, academia, everyone interested) who have more significant influence over the development and future of the technology and its developments. Without the decentralization of AI, democratization will be nearly impossible, and we won't be able to achieve the fruitful interoperable development happening on the internet and one of the most powerful tecnhologies of our time can become completely controlled by a handful of powerful companies in the world.

\subsection{Related Works}

DEAI networks and solutions have been mentioned, introduced, and analyzed in several works, often approached from different angles. For example, some researchers focused only on a single domain, such as healthcare or energy systems, investigated only one kind of ML architecture, such as neural networks (NN), focused only on permissioned systems, or analyzed the possible applications and use cases \cite{tagde2021blockchain, shafay2023blockchain, ai_and_blockchain_integration_ieee}. Furthermore, some reviews, surveys, and systematic literature reviews (SLR) exist that analyze the research in this field with a specific focus. Still, none approaches it from the bottom-up perspective with common building blocks of existing solutions identified.

Liu et al. \cite{survey_ai_ml} explored and presented several benefits of blockchain and ML integration for communication and networking systems, enabling decentralized intelligence, increased privacy and security of ML models, and better data and model sharing. The review paper from Chavali et al. \cite{ai_and_blockchain_integration_ieee} presents several advantages of how blockchain can improve AI and vice-versa. Furthermore, Chavali et al. analyzed multiple projects from industry and gray literature and proposed a DEAI framework, in contrast to our work, where we identify possible building blocks such as service discoverability, service registry, AI services abstraction, marketplace, etc. 

Miglani and Kumar performed a SLR \cite{blockchain_ml_iot} to identify several additional use cases on how blockchain and decentralization can improve AI in Internet of Things (IoT) environments in 5G networks and beyond, including trustless smart contracts, the verifiable open registry for AI, secure access control, and privacy-preserving computation. The positive impact of blockchain technology on AI was also found in the survey paper \cite{wang2021applications}, where decentralized, federated learning, secure data sharing, and decentralized intelligence were researched. There has also been high interest in DEAI from the industry and established companies, highlighting the benefits and advantages of this technology to the existing solutions \cite{hashim2019decrypting}. Nevertheless, to the best of our knowledge, there is no dedicated SLR on general DEAI and its building blocks.

\subsection{Motivation and Contribution}

As already mentioned, most of the current research reviews were focused either on high-level advantages and disadvantages of DEAI or on applying decentralization to a single domain or problem. While the outcomes of those articles provide an overview and the implications of the technology, they do not give an oversight of the common core components of different solutions, while each article addresses and tries to solve some core problem in solitary. None of the papers perform a comprehensive or systematic literature review (from peer-reviewed or/and gray literature), and try to identify building blocks from the bottom-up (ground-up) perspective. Due to this shortcoming of the current state-of-the-art literature, we chose to perform a comprehensive SLR on the topic of DEAI and its building blocks. 

This paper identifies all building blocks of different works and projects from the DEAI field and provides oversight of all the building blocks future researchers and implementers should consider and have in mind when designing DEAI networks, systems, or solutions. If there were several widely used and successful DEAI networks live, we could compare them side-by-side, as it was possible with decentralized storage networks in paper \cite{daniel2022ipfs}, where the following building blocks were identified: network architecture, information security, file handling, file size, and incentivization. Instead, we focus on the comprehensive literature review of published articles, conference papers, book chapters, and other available gray literature.

To the best of our knowledge, the main contributions of this work are as follows:

\begin{itemize}
    \item A comprehensive systematic literature review and analysis of the DEAI field.
    \item The identification of the core DEAI building blocks, which serve as guidelines and components to consider when designing new solutions and networks.
    \item The identification of different metrics to evaluate DEAI systems from different perspectives.
    \item The identification of significant DEAI issues and challenges that must be explored in the future.
\end{itemize}

\subsection{Organization}

The rest of the paper is organized as follows. Section \ref{sec:deai} provides an overview of DEAI and other important AI subfields, diving deep into the field and explaining research directions and open problems. Section \ref{sec:methodology} describes the applied research approach, Systematic Literature Review (SLR), providing the reader with research questions, search strings, searched databases, inclusion and exclusion criteria with limitations, and a description of the screening process. Section \ref{sec:main_contribution} contains the paper's main contribution, which is identifying the building blocks, features, and challenges of DEAI systems and networks, as well as different evaluation methods. In Section \ref{sec:discussion}, we analyze the results, discuss the progress of the field, and provide the learned lessons. The paper is concluded in Section \ref{sec:conclusion}.

\section{An Overview of Decentralized Artificial Intelligence - DEAI}
\label{sec:deai}

This section provides a brief overview and background on the technologies related to the DEAI, emphasizing the differences and roles of different fields - \textit{artificial intelligence}, \textit{distributed artificial intelligence}, \textit{multi-agent systems}, \textit{edge artificial intelligence}, \textit{distributed ledger technology}, \textit{blockchain}, and \textit{decentralized artificial intelligence}.

\subsection{Artificial Intelligence (AI)}

AI is one of the most influential technologies of the 21st century and one of the most active research fields in computer science, with its influences being felt in almost any other research field. The field is concerned with giving computers and machines the ability to solve intelligent tasks and imitate human intelligence \cite{russell2010artificial}. The field of AI can be further split into several subfields, where each area researches different approaches to achieving AI or focuses on various technical aspects. Some of the most known areas are expert systems, symbolic AI, ML, deep learning (DL), artificial general intelligence (AGI), and AI safety \cite{russell2010artificial}. 

The field of AI has been through several periods in history; several AI winters when the development and optimism fell, different subfields becoming popular, such as expert systems in the late 20th century, etc. In the past decade, the most significant breakthrough was achieved in ML and DL, to which better computing resources, such as graphical processing units (GPUs), and the availability of data and datasets (big data), thanks to the internet, became available \cite{krizhevsky2012imagenet}. One of the most significant early breakthroughs was the success of the convolutional neural network (CNN) AlexNET on the image dataset ImageNet for image classification in the year 2012 \cite{krizhevsky2012imagenet}. The next significant breakthrough was the introduction of the concept of attention from the paper Attention is all you need, which led to the fast development and advancements in the large language models (LLMs), which are based on the transformer architecture \cite{vaswani2017attention}. In the past years, ML and DL achieved progress in essentially all of the fields that are focused on different modalities of data or research fields, such as on graphs with graph neural networks (GNNs), time series data with the long-short term (LSTM) neural networks, and others \cite{vaswani2017attention, kipf2016semi, graves2012long}.

Due to the high computing resources and big data demanded by ML and DL models, the training and development of AI ended in the single centralized data centers controlled and most often owned by a single company or entity \cite{montes2019distributed}. Thus, this kind of AI is often called centralized AI (CEAI). This resulted in the development of supercomputers by big tech companies, which also control almost all data on the internet due to owning social networks or critical parts of the internet infrastructure, in turn making them competitive (and in many cases even more successful) to government and public institutions in deploying and developing AI models; thus private sector gaining strong influence on the AI field as whole \cite{stanford_index}.

\subsection{Distributed Artificial Intelligence (DAI), Multi-Agent System (MAS)}

Distributed AI (DAI) is a subfield of AI mainly focused on training (sometimes also inference) AI models \cite{o1996foundations}. Two components of the AI workflows that are distributed most of the time are models and data. This means that parts of the model are distributed across multiple devices (often in different locations) or data is split into various subsets and transferred to multiple devices training the same model. The field started in the 70s and began to be widely developed in the early 90s and was often referred to as DEAI in earlier works, but the latter term was later dropped for different research areas \cite{miiller1990decentralized, chaib1992trends}. While the components of AI can be distributed to different devices and thus are not in a single location as in CEAI, this type of AI is still heavily centralized due to a single entity owning the devices and machines, as well as the whole coordination. DAI enables better and faster computation of high-intensive workloads when single machines cannot handle them since they are efficiently paralyzed across multiple devices. This is very common in the ML and DL field, whereas training of the NN can be faster by several orders of magnitude \cite{gupta2018distributed, chahal2020hitchhiker}.

One of the research directions in the DAI is multi-agent systems (MAS). MAS are systems of autonomous agents that operate without the direct control of human beings and can communicate and cooperate with other agents to solve specific and complex problems \cite{miiller1990decentralized}. The complex problem is often divided into multiple subproblems, whereas each agent is responsible for one issue, and solutions are then constructed from coordination and solutions provided by each agent. While the agents are autonomous and independent of other agents in the system, the systems are still often defined, controlled, and created by single entities or companies to solve their complex problems. Some of the applications of the MAS include computer networking, robotics, smart grids, etc. \cite{mas_survey, jiang2013understanding, nguyen2012agent}.

\subsection{Edge Artificial Intelligence (Edge AI)}

Edge AI is one of the AI fields that emerged in the last couple of years with the fast advancement in IoT devices and their computing capabilities, networking technologies, and big data. The field is also part of the broader field named edge computing, which pushes computation to the network edge and devices, enabling faster response times to the computation workflows and removing traffic from internet highways \cite{zhou2019edge}. The main difference to DAI is that data never leaves the edge devices where the computation is performed, unlike DAI, where the data is split and distributed from a central location, distributed nodes can also be big data centers. Edge AI also enhances data privacy and sovereignty. While model inference (using the AI/ML models) can efficiently run on edge devices, centralized servers still orchestrate training, e.g., federated learning in ML \cite{federated_learning}. Thus, Edge AI can be considered closer to DEAI than DAI but is still under the control of centralized entities.

\subsection{Distributed Ledger Technology (DLT), Blockchain}

Distributed ledger technology (DLT) is the computer science field that is not directly connected to the AI field but is part of distributed systems \cite{rauchs2018distributed}. Researchers of the DLTs focus on solving the complex problems of designing different mechanisms for the nodes in the P2P networks to come to a consensus and single ledger of truth, which can represent financial transactions, data storage information, or any other type of data. The digital ledger can be represented with different data structures, such as blockchain or directed acyclic graph (DAG), and there are different consensus protocol types, such as Nakamoto consensus (e.g., proof of work) and classical voting consensus (e.g., practical byzantine fault tolerance - pBFT).

The most significant breakthrough that also started the broader interest in the field was the invention of Bitcoin in 2009, where the blockchain was introduced by an anonymous computer scientist(s) named Satoshi Nakamoto \cite{nakamoto2008bitcoin}. Bitcoin blockchain was the first digital ledger of transactions and a P2P decentralized network that solved some of the most challenging problems, e.g., double spending. The invention of the blockchain led to major developments in the field over the past decade, with the most notable network being Ethereum, which extended the use cases of the blockchain to general-purpose computation based on the smart contracts paradigm, where blockchain transactions contain the function calls apart from being mere financial transactions \cite{buterin2014next}. The invention of smart contracts led to the development of decentralized applications (dApps), which serve as the underlying foundation of Web3, the new paradigm, and iteration of the World Wide Web (WWW) \cite{belk2022money}. Web3 differs from Web2 in the core principles, whereas the latter focus on efficiency and user experience but sacrifices privacy, autonomy, and self-sovereignty since the technological stack and systems are in complete control of big tech companies (e.g., social networks), Web3 priorities the principles by utilizing decentralized networks, P2P systems, and cryptography. Multiple networks and systems were proposed and implemented for various purposes, such as Filecoin for decentralized storage and The Graph for indexing blockchain data \cite{filecoin, thegraph}. 

\subsection{Decentralized Artificial Intelligence (DEAI)}

As already mentioned, DEAI is a field that intersects the AI and decentralized systems fields, most often (but not strictly) utilizing DLTs and blockchain \cite{montes2019distributed, deai_web3}. While from a technical perspective, some concepts and design choices from DAI and Edge AI are the same or very similar in DEAI, the latter differs, especially in the following objectives: removing a single point of failure of the system, no centralized control of the system, utilizing open-source code and technologies, self-sovereignty of all actors in the system, enhanced privacy, and sharing of the resources (e.g., computing and AI models). 

One example that showcases the mentioned difference can be inspected on the example of federated learning. While federated learning is an example of DAI with a centralized server controlling the training, the centralization can be removed with smart contracts and blockchain (or some other mechanism), which take over the coordination of training \cite{federated_learning, yu2022blockchain}. This approach is called decentralized, federated learning. Another example is MAS, where multiple autonomous agents work together to solve complex problems, while each agent is responsible for a single or several smaller tasks. In the classical approach, a single organization can control multiple agents or even the whole system \cite{dorri2018multi}. In contrast, independent actors run agents in the decentralized approach, and the network they communicate over is decentralized and permissionless \cite{ponomarev2017multi}. In a more generalized way, DEAI is concerned with redesigning the governance, self-sovereignty, access, and control of AI solutions and systems, similar to how other Web3 primitives redesign the WWW, e.g., with digital assets and decentralized identity.

\section{Research Methodology}
\label{sec:methodology}

This paper follows the Systematic Literature Review Guidelines in Software Engineering \cite{Kitchenham2009}. Before designing and conducting the SLR, we wanted to address the following: (1) identify the current status of DEAI architecture and solutions, (2) the common building blocks of these solutions and their possible overlap, and (3) what problems they are trying to address. Based on that, we defined and formularized the following research questions (RQ):

\begin{enumerate}
    \item RQ 1: What is the definition of a DEAI network?
    \item RQ 2: What are the latest proposed architectural designs for DEAI networks?
    \item RQ 3: How can decentralization improve AI and what are the advantages of a DEAI network?
    \item RQ 4: What are the common building blocks of DEAI networks?
\end{enumerate}

The literature search was conducted over seven sources: \textit{ScienceDirect}, \textit{SpringerLink}, \textit{IEEE Xplore Digital Library}, \textit{ACM Digital Library}, \textit{Web of Science}, \textit{Google Scholar}, and gray literature. The latter was primarily found on \textit{Google Search}, with several articles and projects also found through various platforms, e.g., social networks. We first identified the main keywords addressing the field of interest and the RQ stated above. The following keywords were chosen: \textit{\textbf{artificial intelligence}} (AI), \textit{\textbf{machine learning}} (ML), \textit{\textbf{peer-to-peer}} (P2P), \textit{\textbf{multi-agent system}} (MAS), \textit{\textbf{decentralization}}, \textit{\textbf{network}}, \textit{\textbf{architecture}}, \textit{\textbf{platform}}, \textit{\textbf{service}}, \textit{\textbf{marketplace}}. After choosing the relevant keywords we also identified the possible common abbreviation and used them in the search process. We conducted the search on 27 December 2022 using the following standard search string:

\medskip

("p2p ai" \textbf{OR} "p2p artificial intelligence" \textbf{OR} "p2p machine learning" \textbf{OR} "p2p ml" \textbf{OR} "p2p multi-agent system" \textbf{OR} "p2p mas" \textbf{OR} "peer-to-peer ai" \textbf{OR} "peer-to-peer artificial intelligence" \textbf{OR} "peer-to-peer machine learning" \textbf{OR} "peer-to-peer ml" \textbf{OR} "peer-to-peer multi-agent system" \textbf{OR} "peer-to-peer mas" \textbf{OR} "decentralized ai" \textbf{OR} "decentralized artificial intelligence" \textbf{OR} "decentralized machine learning" \textbf{OR} "decentralized ml" \textbf{OR} "decentralized multi-agent system" \textbf{OR} "decentralized mas") \textbf{AND} ("network" \textbf{OR} "architecture" \textbf{OR} "platform" \textbf{OR} "services" \textbf{OR} "marketplace")

\medskip

The following selection and exclusion criteria were defined for the inclusion of articles, with several limitations also imposed to include as many relevant articles as possible for this research topic and field.

\medskip

\textbf{Selection criteria:}

\begin{itemize}
    \item The research addressed P2P and decentralized platform/network/architecture for AI/ML/MAS.
    \item The work proposed a general-purpose solution that was not specific for one problem or domain.
    \item The research was published as a journal article, conference paper, book chapter, or in gray literature.
    \item The work was peer-reviewed, or in the case of gray literature we could identify technical documentation that described the solution in detail.
\end{itemize}

It should be noted that there was no limitation on the publishing date.

\medskip

\textbf{Exclusion criteria:}

\begin{itemize}
    \item The research was not available in the English language.
    \item The full text of the research was not available on the digital library or any of the subscription services.
    \item The research only addressed one application of AI/ML/MAS or proposed application/field-specific network for artificial intelligence models/agents.
    \item The research applied only to distributed AI and not decentralized.
\end{itemize}

\medskip

\textbf{Limitations:}

\begin{itemize}
     \item The research was limited to the six scientific databases/search engines:
     \begin{itemize}
     \item ScienceDirect,
     \item SpringerLink, 
     \item IEEE Xplore Digital Library, 
     \item ACM Digital Library, 
     \item Web of Science, 
     \item Google Scholar, and
     \item Google Search.
     \end{itemize}
    \item The research has to be available before the 27. 12. 2022, when the indexing of potential articles was conducted.
    \item The gray literature was limited to \textit{Google Search}, random detection from different sources, and snowballing.
    \item Results on both Google Scholar and Google Search were limited to the first ten pages per each search query, i.e., equal to one hundred works.
\end{itemize}

\medskip

Due to differences and limitations, e.g., length of the query, between various scientific paper databases and search engines, the above search string was split into multiple search strings resulting in several searches for one source. All search strings are in the Appendix~\ref{app:split_ss}. We then removed the duplicate articles between independent searches and combined all the unique articles for each source to simplify the presentation of the results. 

We screened and inspected all returned articles through two iterations. We checked the paper's title, abstract, and keywords in the first iteration and decided whether to include the article. Table \ref{tab:results_first_iteration} shows the search results. We continued with the second one, in which we read the whole article or finished early in the negative case to decide whether to include the article in the final selection. In the second iteration, we also performed the snowballing process, through which we found several extra articles. Table \ref{tab:results_second_iteration} showcases the number of results in the final selection, i.e., after the second iteration.

\begin{table*}[!ht]
\centering
\caption{The total number of results and number of results after the first iteration per each inspected database.}
\begin{tabular}{cccc}
\hline
\textbf{Database Name}      & \textbf{URL}          & \textbf{Total Results} & \textbf{Results after 1st iteration} \\ \hline
ScienceDirect               & \href{https://www.sciencedirect.com}{sciencedirect.com}     & 781                    & 12                                   \\
SpringerLink                & \href{https://link.springer.com}{link.springer.com}     & 569                    & 15                                   \\
IEEE Xplore Digital Library & \href{https://ieeexplore.ieee.org}{ieeexplore.ieee.org}   & 89                     & 11                                   \\
ACM Digital Library         & \href{https://dl.acm.org}{dl.acm.org}            & 24                     & 0                                    \\
Web of Science              & \href{https://webofscience.com}{webofscience.com}      & 159                    & 11                                   \\
Google Scholar              & \href{https://scholar.google.com}{scholar.google.com}    & 600+                   & 38                                   \\
Gray literature             & \href{https://google.com}{google.com} and others & 250+                   & 12                                   \\ \hline
\textbf{Total}              &                       &    \textbf{2472+}                    & \textbf{99}                         
\end{tabular}
\label{tab:results_first_iteration}
\end{table*}

\begin{table*}[!hb]
\centering
\caption{The number of results after the second iteration, which also included snowballing.}
\begin{tabular}{cccc}
\hline
\textbf{Database Name}      & \textbf{URL}          & \textbf{Results after 1st iteration} & \textbf{Results after 2nd iteration} \\ \hline
ScienceDirect               & \href{https://www.sciencedirect.com}{sciencedirect.com}     & 12                                   & 3                                                        \\
SpringerLink                & \href{https://link.springer.com}{link.springer.com}      & 15                                   & 4                                                       \\
IEEE Xplore Digital Library & \href{https://ieeexplore.ieee.org}{ieeexplore.ieee.org}   & 11                                   & 9                                                        \\
ACM Digital Library         & \href{https://dl.acm.org}{dl.acm.org}             & 0                                    & 0                                                        \\
Web of Science              & \href{https://webofscience.com}{webofscience.com}     & 11                                   & 4                                                        \\
Google Scholar              & \href{https://scholar.google.com}{scholar.google.com}    & 38                                   & 20                                                       \\
Gray literature             & \href{https://google.com}{google.com} and others & 24                                   & 18                                                       \\
Snowballing                 &                       &                                      & 13                                                       \\ \hline
\textbf{Total}              &                       & \textbf{99}                          & \textbf{71}                                             
\end{tabular}
\label{tab:results_second_iteration}
\end{table*}

\begin{figure*}[!ht]
\centering
\includegraphics[width=1\textwidth]{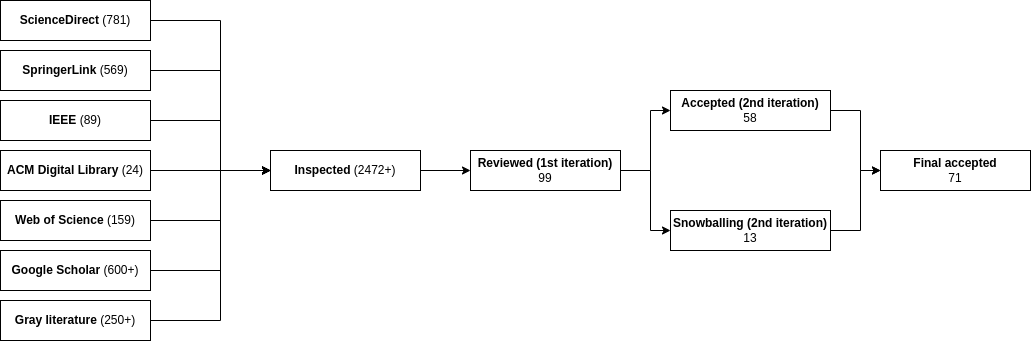}
\caption{Reduction of the papers through the screening process.}
\label{fig:search_results}
\end{figure*}

Most of the results from the first iteration were not relevant to the SLR's objective and were not in line with our inclusion and exclusion criteria. The total number of found results using search queries was over 2472, which was reduced to 99 after the first iteration. After the second iteration, which included reading whole papers and snowballing, we finished with 71 papers investigated and analyzed in detail in the later sections. Figure \ref{fig:search_results} shows the gradual reduction of papers through the screening process.

There were several attributes screened for each paper during the second iteration of the screening process. We focused on some relevant metadata about each paper, such as database, year of publication, place of publication (conference, journal, etc.), as well as other attributes essential to our analysis: \begin{itemize}
    \item building blocks,
    \item features and challenges,
    \item evaluation metrics,
    \item contribution,
    \item supported type of AI models, and
    \item system/network type (public or private/permissioned).
\end{itemize} 

Some building blocks were defined before the research, while most of them were identified during the screening. If a new building block was identified later, we updated the attributes of all previously screened papers. Ultimately, we identified 13 main building blocks presented in Table \ref{tab:build}. Table \ref{tab:attributes_screening} shows the attributes that were screened. Furthermore, we analyzed each paper based on these additional attributes, such as if the work proposed architecture, concept or was just a review paper.

\begin{table}[!ht]
\caption{Building blocks, features, and challenges identified in the research.}
\begin{tabular}{ p{4cm} p{4cm} }
\hline
\textbf{Building block}     & \textbf{Description}                                                    \\ \hline
Registry                 & There exists the trusted registry that contains the list of all available AI models.                                                                                                                                                \\
Incentivization          & Users pay for using AI models, which results in the incentives for publishers of AI models.                                                                                                                                         \\
Marketplace              & Open marketplace for trading AI models. This building block is tightly tied to the ownership.                                                                                                                                       \\
Reputation               & Reputation systems enable the ranking of the AI models and help users identify good and bad actors in the system.                                                                                                                     \\

Ontology                 & AI models can be described by the well-defined ontology, enhancing interoperability, standards, and AI model discoverability.                                                                                                             \\
Discoverability                & There exists a system or language that can efficiently find needed AI models, e.g., models that can classify birds on night images.                                                                                             \\
Training (computation)   & The system supports AI agent/model training.                                                                                                                                                                                        \\
Inference (computation)  & The system supports AI agent/model inference.                                                                                                                                                                                       \\
Ownership                & There is a transparent provenance of who owns the AI model (through time). The owner of the model can change due to various reasons.                                                                                                    \\
Data                     & System provides mechanisms for handling data, i.e., data management, and uploading data to AI agents/models.                                                                                                                            \\
Governance               & There exist strict rules of who makes the decisions in the decentralized system, and who governs the future evolution of the system.                                                                                                \\
Cryptography and Privacy & Cryptography enables several features in the system: verifiability of actions of AI agents (validity AI), privacy, identity ...                                                                                                     \\
Identity                 & Each AI model/agent can be identified by a globally unique identifier and have its own identity.                                                                                                                                      \\
\hline
\textbf{Feature}     & \textbf{Description}                                                    \\ \hline \\

Standards and Specifications                & The systems support well-defined and used technical standards, resulting in interoperability between different frameworks as well as between different AI models, e.g., connecting the output of one model to the input of different models. \\

Security                 & The system is secured against different attack vectors, such as denial-of-service (DoS) attacks.                                                                                                                                    \\

Transparency             & System provides an exact overview of AI lifecycle and how the model changed over time. \\

\hline

\textbf{Challenge}     & \textbf{Description}                                                    \\ \hline

Updating global models   & Models can be improved over time (for example when trained on new data) and DEAI enables updating the existing models. Updating the models can be incentivized.                                                    \\

\end{tabular}
\label{tab:build}
\end{table}

\begin{table*}[!ht]
\centering
\caption{Attributes retrieved for each work during screening.}
\begin{tabular}{ccccccc}
\textbf{Place of publication}                                                                                                                  & \textbf{Contribution}                                                                       & \textbf{Database}                                                                                                                                                                   & \textbf{Year} &  \textbf{Evaluation metrics}                                                                                   & \textbf{Type of AI models}                                                                  & \textbf{Network type}                                     \\ \hline
\begin{tabular}[c]{@{}c@{}}A - journal\\ B - gray literature\\ C - conference\\ D - paper\\ E - thesis\\ F - book chapter\end{tabular} & \begin{tabular}[c]{@{}c@{}}A - architecture\\ B - concept\\ C - review\end{tabular} & \begin{tabular}[c]{@{}c@{}}A - ScienceDirect\\ B - SpringerLink\\ C - IEEE\\ D - ACM\\ E - Web of Science\\ F - Google Scholar\\ G - gray literature\\ H - snowballing\end{tabular} & 1990 - 2022   & \begin{tabular}[c]{@{}c@{}}A - read latency\\ B - write latency\\ C - extensible for users\\ ...\end{tabular} & \begin{tabular}[c]{@{}c@{}}A - AI models\\ B - ML models\\ C - DL models\\ ...\end{tabular} & \begin{tabular}[c]{@{}c@{}}A - private\\ B - public\end{tabular}
\end{tabular}
\label{tab:attributes_screening}
\end{table*}

\section{Building Blocks, Features, and Evaluation Methods}
\label{sec:main_contribution}

This section contains the main contribution of this paper, based on the methodology defined in the previous section. We define the building blocks, features, challenges, and evaluation methods of the DEAI networks and systems. We hope future research and projects will focus on and investigate how these blocks fit into their solutions. In the following paragraphs, we use terms like AI service, agent, and models interchangeably to stay on the high level and not focus on too much technical terminology. While some papers differentiate between these terms, they have similar meanings in various works. Most of the building blocks are relevant to all subtypes of AI, i.e., machine learning, deep learning, and symbolic systems, although some are more focused (but not strictly) on one subtype. In the latter case, this is mentioned in the building block's description in this chapter, e.g., training (computation) for ML models. All articles with their identified building block are found in Appendix~\ref{app:all_articles}, and Figure~\ref{fig:bb_connections} shows the connection between building blocks. Some building blocks are not connected to others since they are not strictly required for others but can still play a role, e.g., governance in the marketplace.

\begin{figure}[!ht]
\centering
\includegraphics[width=1\columnwidth]{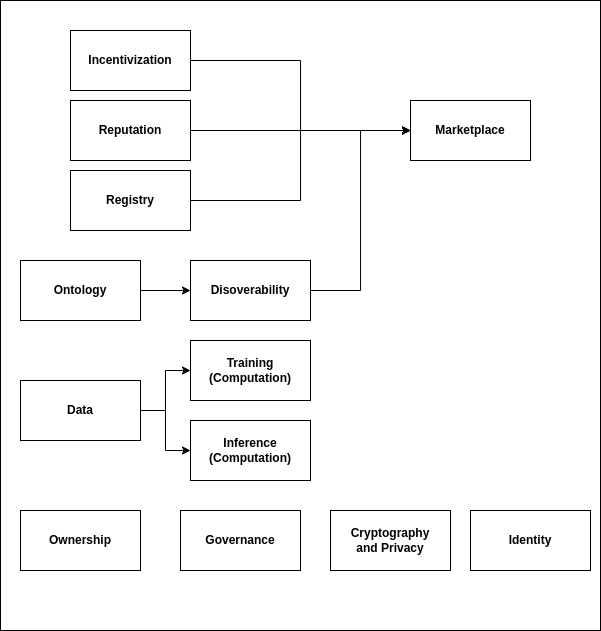}
\caption{Connections between building blocks.}
\label{fig:bb_connections}
\end{figure}

\subsection{Building Blocks}

This subsection presents the list of identified building blocks. With term building blocks we group and reflect the critical components and processes, which cover a basic or critical feature of a DEAI network. Each of the building blocks covers one of the major, basic, or critical parts of the DEAI mosaic, whereby each one of those is singular and self-sufficient, as well as modular and connectable to others. 

\subsubsection{Registry}

One of the biggest hurdles in the AI space is the distribution of AI models once they are ready to be used. Due to the CEAI being closed from the outside participants and no standardized way for the distribution process, researchers and companies do a lot of double work and cannot easily use the new models as they are created. In reality, most of the work stays in the specific papers, where results are often hard to repeat, or in GitHub repositories, which require specialized knowledge to set up the environment and use the models. There has been improvement in the last period with several companies, such as Hugging Face \cite{hugging_face_spaces}, introducing platforms to distribute and share AI models with users, significantly increasing the models' usability. But these systems are closed, run behind walled gardens, and are controlled by a single company or entity, making them susceptible to a single point of failure. Because the registries are an integral part of their business models, they are often curated so that users are tied to their single ecosystem. One of the most challenging problems for the registries is the curation of the entries, which often becomes problematic with centralized systems. Therefore, there is a need for an open and permissionless registry (or also index) for all AI models \cite{blockchain_ml_iot}.

Various types and implementations of registries were defined and introduced in different analyzed works. In multiple research works and projects, registries play a central role in the solution and contain metadata about available AI services and metadata about their organizations and developers. The code for the registries, i.e., rules and associated actions (e.g., updating entries), is often published on a permissionless system and networks, such as blockchain or other DLTs. But the metadata itself is usually stored on the decentralized storage networks, such as IPFS, which can handle larger data flows, thus containing more extensive and more comprehensive descriptions of services \cite{singularitynet, dinemmo, fabricfl, trustless_api, masurkar2019decentralized, hackathon_ai}. One of the frequent attributes is the version of the model, which makes it possible to have a transparent and verifiable chain of the model updates throughout history \cite{adel2022decentralizing}. Aside from registries about AI models, there is comprehensive research on permissionless and open data registries, which often play a central role in AI workflows (especially in the ML field) \cite{ocean_protocol, proof_of_learning}. While different types of registries have common characteristics, they still differ in various aspects since the primitives they present, i.e., AI models and data, are different. Project Fetch.ai is developing an agent explorer, which makes the discovery, exploring, and finding of all the autonomous economic agents (AEA) on their network straightforward \cite{fetch.ai}. Algovera AI is developing Metahub based on the Ocean Protocol to aggregate all AI assets registered on various blockchain networks \cite{blythman2022libraries}.

\subsubsection{Incentivization}

One of the most challenging goals of global decentralized and P2P networks is designing them for an extended time. Because a profitable company does not run them, they must attract network maintainers, i.e., in the case of a blockchain network those are node operators. These are compensated through various token incentives mechanisms. One of the significant breakthroughs in incentive mechanisms was the development of non-fungible tokens (NFTs), which bootstrapped the digital economy of intellectual property (IP) works and royalties for creators \cite{wang2021non}. An incentive mechanism should be built into the network design itself. Otherwise, this has to be developed outside the system for sustainability, resulting in multiple ad-hoc solutions that worsen the system's user experience. 

The most straightforward incentive mechanism in the DEAI system is paying for consuming AI services, whereas users pay the developers of the AI models for their usage. In multiple works, models were designed and proposed to be deployed on-chain as smart contracts, chain codes, or hosted and run by the network nodes in the AI-specific blockchain \cite{adel2022decentralizing, deepbrainchain, oraichain, platon, giza, cortex}. While this approach focuses on running AI models directly by node operators, the alternative solution is to use on-chain mechanisms to register, discover, and facilitate the payment process with the execution happening on the external third-party network or server \cite{singularitynet, hackathon_ai, fetch.ai, dinemmo, sketching, trustless_api, lu2018enabling, ponomarev2017multi}. The latter approach often includes state channels, user token deposits, and other transaction payments. Ocean Protocol is working on assets for data ownership, and their utility is providing data access, while Algovera AI is developing NFT-based access to AI services \cite{ocean_protocol, algovera_ai}.

Researchers also proposed incentive mechanisms for the compensation of the model training. These incentives are being again the tokens or other payment options, and often also access to the best globally trained model (or improved version of the local one). This approach can be called decentralized collaborative training. Authors proposed mechanisms using smart contracts for compensating users for sharing the data or sending model updates, e.g., gradient updates \cite{micro_coll, masurkar2019decentralized, nguyen2021marketplace, johnson2021ichain}. Other works proposed compensation rewards for directly improving and making the model better \cite{learning_markets, liang2021omnilytics, numerai, kurtulmus2018trustless}. Decentralized federated learning with reward mechanisms was proposed in several works, where users participate in the training during the whole process (not just sending partial results) and are compensated for offering the device computation power \cite{weng2019deepchain, pds2, kim2019efficient, fabricfl, yu2022blockchain}. Other P2P learning mechanisms were introduced where participants directly collaborate with peers in the P2P network and work together to improve the AI models \cite{rao2020bittensor, boubouh2020robust}. Another proposed solution is the development of a new blockchain network with a consensus mechanism specifically designed for training the AI models \cite{kusi2020training, teerapittayanon2019daimon, li2020blockchain, proof_of_learning}. 

\subsubsection{Marketplace}

Related to other building blocks, such as registry and incentivization, marketplaces offer a platform to monetize AI models on a scale. AI marketplaces are defined in literature as a place that enables AI developers to monetize their models. The following roles most often appear in the works (but not all are strictly necessary): AI developers, data owners, end-users, AI model auditors, and cloud or computing vendors. AI developers want to monetize their developed models, data owners want to monetize their data that can be used for model training, end-users want to use services for different purposes, auditors are responsible for auditing the services and AI models, and computing vendors provide infrastructure for training and inference of the models. Apart from AI marketplaces, many works focused only on data or general-purpose computing resource marketplaces.

Authors in \cite{sketching} define a high-level definition of AI marketplaces and analyze them from technological, economic, and regulatory perspectives. Multiple projects proposed or are implementing a variant of AI marketplaces that offers monetization mechanisms for developers and users to use the services by paying with cryptocurrency \cite{singularitynet, fetch.ai, hackathon_ai, dinemmo, blythman2022libraries, oraichain, effect_network, cortex}. SingularityNET integrated support for state channels to facilitate payments between service providers and users to lower the fees associated with Ethereum transactions \cite{singularitynet}. Fetch.ai is developing a platform for AEA and enabling users to pay agents to perform tasks on their behalf \cite{fetch.ai}. Researchers explored marketplaces for federated learning, where computing vendors contribute resources to models listed on the marketplace \cite{nguyen2021marketplace, learning_markets}. \cite{pds2, somy2019ownership, liang2021omnilytics} explored data marketplaces for data owners. Works \cite{platon, nunet, ponomarev2017multi} explored marketplaces for computation resources that AI models and services utilize. Authors in \cite{trustless_api} proposed utilizing blockchain to enable generic trustless API marketplaces, with ML being mentioned as one of the possible use cases.

\subsubsection{Reputation}

Anonymity and pseudonymity are one of Web3's most significant advantages and helped bootstrap multiple ecosystems that rely on those features. While they have several advantages, there are also some drawbacks, such as Sybil attacks and trustworthiness in identities and data \cite{douceur2002sybil}. That is also especially hard in decentralized systems where no single entity can verify users through various tools, such as Know Your Customer (KYC). Therefore, there is a need for more advanced mechanisms that must be implemented to determine the goodness, trustworthiness, and reputation of the actors in the systems, i.e., models, developers, users, and others.

Project Bittensor is developing a P2P internet-scale NN where peers learn the ranking of their neighbors based on contributions, and these scores are scored on the blockchain with high-ranking peers receiving higher rewards \cite{rao2020bittensor}. Authors researching decentralized, federated learning proposed the Krum algorithm and its many variants to design systems to be immune to bad actors and distribute rewards based on contribution and credibility \cite{blanchard2017machine, peyvandi2022privacy, yu2022blockchain}. Many works proposed P2P reviewing of AI models and calculating reputation based on that, similar to online e-commerce platforms, while others argue that in addition to that, several mechanisms must be put in place to avoid cheating the system dynamics, e.g., model assessments by experts \cite{teerapittayanon2019daimon, sketching}. Authors in \cite{platon, oraichain, effect_network, numerai} proposed rating AI services according to the evaluation score on dedicated datasets and building a reputation scoring system on top of that, with additional mechanisms, such as token slashing in case of malicious acts, calculating average metric over a longer period, or comparing results by utilizing multi-party problem-solving. Project SingluarityNET is working on a proof of reputation based on liquid rank and democracy to replace a simple voting system. AI agents rank other agents, and reputation is directly integrated into governance, influence in the system, and system rewards \cite{singularitynet}.

\subsubsection{Ontology}

In information science, ontologies add formal representation and naming conventions for the otherwise seemingly non-structured entities, enhancing the understanding of properties and relations between entities \cite{vickery1997ontologies}. Related to the AI field and agents, this translates to defining a way to describe their goals, essence, place in the systems, jobs, and other characteristics that could improve defining and creating relations between agents. While the advantages of using ontologies remind to some degree of standards, standards are more concerned with lower-level technical specifications, and ontologies are with higher-level descriptions and types. Both contribute and help define order in the DEAI system and improve interoperability.

There is comprehensive research on ontologies and agents in the MA research field, with researchers proposing different methodologies for ontology-based agent system development (MOBMAS)  \cite{tran2008mobmas}. These methodologies and frameworks mainly were focused on MAS organization designs, internal agent design, agent interaction designs, and complete architectural design, focusing on P2P implementations \cite{tran2008mobmas, tran2008preliminary, tran2007design}. Authors in \cite{gorodetskii2008development} explored adding ontology to the P2P agent implementation based on the FIPA Nomadic Agents Working Group. Project SingularityNET is working on its own AI-DSL that includes hierarchical data and agent ontology, focusing on standardizing input and output structures of the agents, the time and costs required to perform agent's tasks, financial and payment information, and building high-level end-user language for defining all that information \cite{montes2019distributed}. Effect Network is developing on the decentralized registry of AI services enhanced with rich ontology and defining technical schema for inputs and outputs \cite{effect_network}. Authors in \cite{pds2} emphasize the challenge of machine-understandable semantic metadata for data and model discovery and how both data marketplaces and decentralized machine learning can benefit from using ontologies.

\subsubsection{Discoverability}

While this building block is highly connected and builds upon the other ones, i.e., ontology, it's still mentioned separately due to relying on additional advanced mechanisms apart from other building blocks that can still be incorporated standalone. Model discoverability refers to more advanced strategies for finding agents in the system and not relying on simple query parameters like the registry. It enables users - humans and other AI agents - to find services they need and connect to them entirely programmatically.

Multiple works proposed various discoverability mechanisms in their solutions, incorporating agent metadata, specific domain attributes, extended model descriptions, and other attributes \cite{hackathon_ai, singularitynet, dinemmo, fetch.ai}. As mentioned before, the project SingularityNET is working on AI-DSL for better model discoverability as well \cite{singularitynet}. Effect Network is also exploring advanced model discoverability approaches, which according to them, should increase collaborations between AI agents and services \cite{effect_network}. Several MAS frameworks included components that serve as a module for discovering other agents in the P2P networks, often specialized for specific tasks \cite{gorodetskii2008development}. PlatON Network developers argue that the discovery and recommendation of AI services can be improved by considering the reputation and other scoring systems in the network, as well as adding ML models that learn on historical data \cite{platon}. Authors in \cite{pds2} proposed similar approaches for data discovery based on ontologies. Several alternative discovery mechanisms were presented in various works, such as \cite{ponomarev2017multi}, where the task description is registered on the blockchain (or any other suitable P2P network), while AI agents are listening for events and searching for available jobs that are relevant to them. \cite{sketching} proposed a system where users describe the problem, and the AI marketplace autonomously finds matching companies that can solve that problem.

\subsubsection{Training (Computation)}

The most popular AI subtype is ML, whose workflows consist of two main stages: training and inference. While the latter stage is often more decentralized than the former since it is much less computationally intensive and can be performed on consumer devices, both steps are usually performed entirely centrally (CEAI). Two significant factors are the need for powerful hardware and a lot of data, especially for larger ML models such as LLM. Several solutions in the literature focus on decentralizing both stages, but some focus only on training or inference due to different requirements. Thus, both stages were split and analyzed as separate building blocks.

Training models (especially the ones used in production) often require high computing and power resources. Many works proposed introducing a separate computing layer (or network) to the DEAI one, which serves general-purpose computation tasks, but is specialized and be used for training AI models \cite{singularitynet, nunet, filecoin, bacalhau, effect_network, platon, deepbrainchain}. These systems most often incorporate payment mechanisms in the form of cryptocurrency. Authors in \cite{goel2019deepring} proposed a new blockchain system designed specifically for training deep neural networks (DNNs). Several works proposed new consensus algorithms focused on training AI models \cite{proof_of_learning, kuo2018modelchain, li2020blockchain, du2022accelerating, teerapittayanon2019daimon, kusi2020training, rao2020bittensor, weng2019deepchain}. Researchers also proposed model training in a P2P manner without using blockchain or other DLTs, with often non-financial incentives, such as access to training and final globally trained model \cite{colink, boubouh2020robust}. Decentralized training can also be achieved using smart contracts and other programmability mechanisms on existing blockchain platforms, which can perform gradient aggregation, gradient updates, a variant of federated learning, and other advanced ways to train models jointly \cite{fabricfl, zhou2020pirate, zhang2020sablockfl, majeed2019flchain, kurtulmus2018trustless, yu2022blockchain, peyvandi2022privacy, kim2019efficient, liang2021omnilytics, dinemmo, micro_coll}.

\subsubsection{Inference (Computation)}

The second step is inference, i.e., using ML models in the production with new, often unseen data. The inference is less computationally intensive than training as it requires one forward pass of the input data; thus, the solutions often differ from those proposed for training. But as it turns out, many decentralized training solutions also offer inference capability (but usually not the other way around).

DEAI strives for openness as much as possible and open-source solutions and models; therefore, most systems focus on public models accessible to everyone. But some solutions, such as AI marketplaces in many variants, support public and private models because the inference is often executed directly on the computers of model providers \cite{fetch.ai, singularitynet, sketching, adel2022decentralizing}. Authors in \cite{ponomarev2017multi, lu2018enabling} proposed a system where inference tasks are published on the blockchain or P2P networks, and model providers can register for the job, execute the AI algorithm locally and provide the results to the system. Similar to training, multiple authors proposed special purpose P2P (blockchain) networks where node operators run inference for users \cite{oraichain, platon, cortex, deepbrainchain, trustless_api}. Another approach to decentralized training is deploying AI models in the form of smart contracts (or L2 programs), enabling them to be callable from any other smart contract in the system \cite{giza, micro_coll}. Several projects, such as Filecoin and related The Bacalhau Project, started as decentralized storage networks and are now transitioning to more generalized compute-over-data systems, a way to enable implementing programs that perform AI inference over data  \cite{filecoin, bacalhau}.

\subsubsection{Ownership}

Ownership is one of the features that is being redefined and is at the center of the Web3 movement and permissionless systems in general. It refers to the rules about the digital primitives, e.g., assets or data, that are defined in the systems and control who can perform specific actions. For example, in social networks: in Web2, the user and platform owner can update, delete, and post posts in the name of a user, while in Web3, only the user can perform these actions. The adoption of financial assets, such as cryptocurrencies and NFTs, focuses a lot on ownership and who, in reality, controls what they own. While the latter adoption is already underway, the ownership of data and AI agents is still being researched and is in the experimental phase due to the different characteristics of digital primitives. But in the majority of cases, the ownership is moving in the direction that the owner of the asset is the one who controls the underlying cryptographic keys.

Much research and work has been done on the data ownership problem, where one of the goals is to remove a single gatekeeper of the data that can prevent users from accessing their data while still preserving a high degree of data privacy. This can be achieved by utilizing approaches based on the blockchain, DLTs, or other P2P networks where a distributed set of nodes serves as a gateway to data and cryptography (digital signatures, encryption) serves as ownership and privacy layer \cite{pds2, platon, ocean_protocol, clifton2022decentralized, sketching, kumar2020marketplace, wang2021applications}. Digital ownership can also be in the form of NFTs, which represent the intellectual property (IP) of data, or other cryptographically verifiable digital primitives, such as Decentralized Identifiers (DIDs) and Verifiable Credentials (VCs) \cite{ocean_protocol, blythman2022libraries}. Authors in \cite{somy2019ownership} divided ownership based on the type of entities in the DEAI system, i.e., data owners, model owners, and cloud/computing owners. Several advanced techniques, such as data splitting and replication techniques, were proposed for data so that only the owner has complete control and overview of all data \cite{somy2019ownership, trustless_api}. Projects SingularityNET, Fetch.ai, Giza, and others proposed direct ownership of AI services and agents by making the ownership rights transferable \cite{singularitynet, fetch.ai, giza, ai_bl_integration}.

\subsubsection{Data}

Data is often considered one of the most essential components of AI systems since it serves as the knowledge base for agents and enables them to learn to make intelligent decisions in the future. This is especially evident in the ML field, which requires vast amounts of data to achieve high accuracy and internet-enabled data sharing and collecting more easily. While at the beginning of the AI research, researchers were not concerned much about data privacy, ownership, and other property rights, as seen with the current IP problems with LLMs, the conversation around these topics has been increasing in the past few years. Based on the reviewed literature and projects, the data component doesn't have to be strictly part of the DEAI system. Nevertheless, it is highly interconnected, and many works include it or provide possible solutions and recommendations. 

Federated learning is one of the approaches highly concerned with data privacy, with data never leaving the devices of the data owners or contributors, but only gradient updates are sent from devices to the server that aggregates all the updates and performs the training iteration of the model. Decentralized federated learning goes one step forward by eliminating and replacing the centralized servers with the decentralized alternatives \cite{zhang2020sablockfl, peyvandi2022privacy, majeed2019flchain, yu2022blockchain}. Another data privacy preserving technique that has been explored a lot lately is (fully) homomorphic encryption (FHE), a method where computation is performed over encrypted data, as seen in \cite{gentry2009fully, pds2}. Projects Ocean Protocol, Filecoin, and IPFS working on data and access decentralization are planning to add computing capabilities and foundations for DEAI \cite{ocean_protocol, filecoin, ipfs}. Several works proposed approaches for storing encrypted data on the blockchain or DLT, with smart contracts serving as a gateway and access policy mechanism \cite{privacy_svm, trusted_collaborations}. A lot of research has also been done on ML datasets and their splitting (training, validation, and test), where the training dataset is being revealed, but the test dataset is hidden or revealed in the latest stages of the ML workflows \cite{numerai, kurtulmus2018trustless, teerapittayanon2019daimon}. Authors in \cite{blythman2022libraries} proposed extending the existing AI and ML framework with capabilities to work with data on decentralized storage networks, such as IPFS. \cite{micro_coll} proposed a protocol where users collectively construct datasets by uploading data for the AI models and reward through smart contracts.

\subsubsection{Governance}

Governance is one of the building blocks that is not very technical but refers to the social year of the DEAI system. Nevertheless, it is one of the most critical components of any decentralized system since there needs to be a foundation put in place without a single entity in control of the whole system - development, maintenance, and any decision-making. There has been comprehensive research on the governance in blockchains and other DLTs, with different approaches, such as on-chain governance, off-chain development, and off-chain governance, analyzed \cite{pelt2021defining}. Governance touches other building blocks, such as curation of the registries and setting parameters for incentivization mechanism, and other parts, such as funding the research and development activities \cite{clifton2022decentralized}.

Several works proposed and researched various governance mechanisms. Project Oraichain, Fetch.ai, PlatON Network, and other projects from Web3 ecosystems implemented governance decision-making based on their project's utility tokens \cite{oraichain, fetch.ai, platon}. While this approach is widely used and popularized by many dApps and similar kinds of projects as well, there are many well-known issues such as the plutocracy problem and Sybil attacks \cite{kervsivc2022using, clifton2022decentralized}. One of the possible approaches includes forming a decentralized autonomous organization (DAO) overseeing the system, whose members can vote on important decisions, with the voting power of members being defined by a particular reputation-scoring function. Project SingularityNET is researching proof of contribution and liquid democracy, while Ocean Protocol proposed various forms of AI-governed DAOs \cite{montes2019distributed, trent_ai_daos}. Authors in  \cite{adel2022decentralizing} proposed the self-governance of AI models directly by their creators. \cite{sketching} analyzed and investigated the importance and influence of current and possible future regulations by world nations that AI marketplaces will have to follow and should be considered in the governance process.

\subsubsection{Cryptography and Privacy}

Cryptography is one of the most important building blocks of the current generation of decentralized systems and many other digital technologies. It plays a major role where privacy, verifiability, and security are critical parts of the system. In the Web3 space, cryptography, more precisely public key cryptography, serves as the basis for proof of ownership of all assets, whereas users prove their ownership with digital signatures. Encryption also plays a significant role in preserving users' privacy and data. Advanced cryptographic techniques are also core blocks of blockchain scalability solutions, such as validity proofs for L2s.

One of the most active fields in the DEAI is ZKML, also called validity ML, which focuses on the verifiability of the ML computation, making the ML computations provable and verifiable. Because these zero-knowledge proofs (ZKPs) are still very computationally intensive today, they can be used only for inference and not training \cite{ezkl, giza}. Authors in  \cite{singh2022zero} proposed and explored a verifiable computation approach with ZKPs for every step of the standard ML pipeline. Apart from verifiability, encryption is the foundation for trusted and secure data exchange between network participants, enabling only certain entities to access the actual data \cite{ocean_protocol, peyvandi2022privacy, learning_markets}. Paper \cite{pds2} proposed an approach based on FHE, enabling computation to be performed on the encrypted data and operations executed in the TEEs like Intel SGX. While FHE works and is very useful for several use cases in theory, there are still numerous challenges and complexity problems before it can be used in production for medium and larger AI models. While federated learning itself improves privacy by enabling ML models to be trained over multiple devices without data sharing, cryptographic techniques in decentralized, federated learning can also preserve privacy when sharing the gradients using various methods like differential privacy algorithms \cite{colink, zhang2020sablockfl, peyvandi2022privacy, kim2019efficient}. Some works proposed proof of execution, proof of improvement, verification proofs, and cryptographic approaches for auditability by employing different kinds of commitments \cite{oraichain, teerapittayanon2019daimon, weng2019deepchain}.

\subsubsection{Identity}

Digital identity enables entities to be uniquely identified in the digital systems, enabling trustiness in communication and establishing the source of trust and truth between otherwise unidentified and (pseudo)anonymous actors. While digital identity was most often associated only with human beings, it also provides a comprehensive tool for identifying services, datasets, AI agents, organizations, regulators, etc.

Project Fetch.ai is researching using decentralized identity primitives, such as DIDs and VCs, for communication between AI agents over DIDComm and issuing different types of credentials between agents \cite{fetch.ai}. Ocean Protocol utilizes DIDs for the dataset identification, containing metadata such as service endpoints and data checksum \cite{ocean_protocol}. SingularityNET integrated identification information for organizations that register AI services in their network and metadata for AI services, e.g., description, endpoints, and pricing information \cite{singularitynet}. Project OpenMined is working on integrating DIDComm in federated deep learning use cases, where VCs provide a way to establish trust between actors that participate in the decentralized training and enable the certification of AI researchers by different regulators \cite{openmined_identity}.

\subsection{Features}

Throughout the SLR, we also identified several features. These are not standalone building blocks with strict boundaries incorporated into the solution but should be considered when designing and working on previously described building blocks.

\subsubsection{Standards and Specifications}

One of the main drivers behind the network effect of the Web3 ecosystem and the ability to combine different components and applications as "Lego bricks" are standards and guidelines that greatly enhance the interoperability between products developed by independent entities \cite{katz1994systems}. This can be significantly seen in the Ethereum community with Ethereum Improvement Proposals (EIPs) and Ethereum Request for Comments (ERCs), including ERC20 for cryptocurrencies and ERC721 for NFTs. By strictly following and building on top of the standards, platforms can integrate digital primitives quickly, improving the development and user experience and maintaining high interoperability. One example is ERC20 crypto assets and decentralized exchange (DEX) in the decentralized finance (DeFi) field. Related to DEAI, there are opportunities to define strict and extensible standards for many primitives, such as datasets and their semantic meaning, using linked data, or AI agents, which would improve their interoperability and composability.

Standards (that also help with model discoverability) were mentioned as one of the core components of the DEAI system in several works  \cite{ai_bl_integration, deai_edge}. Many projects are researching different ways to support standards and define rules on how to use them in their platforms, such as SingularityNET, with developing AI domain-specific language (AI-DSL), and Ocean Protocol, which is focused on datasets and ways to describe them with semantically linked data \cite{singularitynet, ocean_protocol}. Projects that are part of the zero-knowledge machine learning (ZKML) field, like Giza and ezkl, are supporting NN format Open Neural Network Exchange (ONNX) during ZK-SNARKS generation and are working to make developing NNs easier in multiple frameworks \cite{giza, ezkl}. Authors in \cite{learning_markets, unified_framework, trusted_collaborations} proposed and included in their works guidelines for following command standards and defining strict policies in the marketplaces, frameworks, and other types of platforms for reuse and interoperability purposes. The lack of focus on standards was also highlighted in work \cite{blythman2022libraries}.

\subsubsection{Security}

Decentralized systems have many advantages over centralized solutions, but several new security concerns must be considered when designing such systems. That also holds for DEAI systems, which networks and solutions must deal with problems and attacks such as malicious node operators, Sybil attacks, and distributed denial-of-service (DDoS). Therefore, solutions must include unique mechanisms to mitigate them and decrease the possibility of them happening.

One of the most adopted approaches to prevent node operators from behaving maliciously lies in crypto-economics, i.e., token staking and slashing in the case of malicious actions. The use of trusted execution environments (TEE) is also often recommended for secure computations and other security considerations \cite{sabt2015trusted}. While this usually solves the problem for network operators, other techniques must be implemented for AI developers and users of the system. Authors in \cite{learning_markets} proposed token deposit and credit management mechanisms to prevent DDoS and Sybil attacks, with credit score rankings being permanently and uniquely linked to the user's accounts. Project DeepBrainChain proposed a multi-layer network with message routing and relayed nodes for DDoS protection\cite{deepbrainchain}.

\subsubsection{Transparency}

This building block differs from others in that it's not something to include in the system but is a feature and result of using the building blocks mentioned before and other related technologies. It is one of the essential characteristics of decentralized networks, such as blockchain and other DLTs, which increases the trustworthiness of data. For several use cases, the most obvious financial ones are, being public, always available, and cryptographically verifiable data, and information improves trustiness and efficiency, especially for end-users. Using different levels of transparency, i.e., fully transparent or semi-transparent, can also help with many emerging AI regulations and, in general, brings several benefits with verifiable models and agents and increased trust.

\subsection{Challenges}

Similar to the features introduced in the previous section, challenges are not standalone building blocks and are something that authors should consider when designing certain building blocks. We identified one challenge that repeatedly appeared throughout several works and solutions.

\subsubsection{Updating Global Models}

For every dataset and related task, there likely exists a model that performs the best for a particular metric. For example, there is a model that performs best in detecting birds in nighttime images, detecting sarcasm in Tweets in English, etc. This building block is connected to training, which can result in the best global model when trained on diverse datasets of participants, and registry, which can be structured to filter models for a particular task.

Multiple approaches for the global distribution of the best AI models were proposed. Several works implemented systems based on smart contracts, which update the models when a certain metric is improved and offer incentives for that \cite{kurtulmus2018trustless, learning_markets, dinemmo, micro_coll}. Decentralized federated learning based on smart contracts, when it doesn't include special privacy-preserving protection techniques, results in a publicly available global model \cite{majeed2019flchain, fabricfl, openmined_identity, zhang2020sablockfl}. Projects and works that proposed an AI-specific blockchain network and consensus often also contain the registry containing the best models obtained through training \cite{proof_of_learning, zhou2020pirate, li2020blockchain, du2022accelerating, teerapittayanon2019daimon, weng2019deepchain, adel2022decentralizing}. Project Numerai implemented a competition-style approach to reward users for contributing the best model for a specific dataset and task \cite{numerai}.

\subsection{Evaluation Methods}

Throughout the review of all identified articles, we also screened approaches through which the authors evaluated their method. This is one of the most important aspects because different approaches are often hard to compare without hard-defined evaluation approaches. While the proposed solution was only presented and introduced in many cases with some sort of qualitative analysis, e.g., increased privacy, removal of a single point of failure, and transparency, several works included details of the more measurable evaluation methods. We focused on the quantitative techniques, where the authors produced measurable data and performed the experiment.

While some evaluation methods differ entirely from others and have little in common, multiple can be grouped due to specific similarities, e.g., methods related to blockchain networks. The latter methods include measuring block times and comparing operations throughput, calculating gas usage and costs when using smart contracts, state of decentralization of the blockchain networks (e.g., number of nodes), how many transactions per second (TPS) blockchain can process, and others \cite{micro_coll, ezkl, somy2019ownership, kim2019efficient, learning_markets, trusted_collaborations, nguyen2021marketplace, du2022accelerating}. Other evaluation methods are concerned with the scalability perspective and AI models, i.e., how many models can the system handle, how many models can be freely added by anyone, support for different modalities and types of models, how many models can be trained per second, can new users join the system or only selected ones, etc. \cite{adel2022decentralizing}. Several works focused on particular training or inference metrics, like time and different metrics, and compared the centralized and decentralized approaches \cite{numerai, weng2019deepchain, privacy_svm, goel2019deepring, kusi2020training, zhang2020sablockfl, li2020blockchain}. For specific subfields, like ZKML, one of the most critical metrics is cipher and proof sizes \cite{giza, ezkl, weng2019deepchain}. Projects and research works, which were in addition to AI, also focused on data use cases, storage consumption, retrieval times, and other metrics that are essential for measuring the performance of the proposed systems \cite{zhou2020pirate, filecoin}. One of the metrics that best showcase the advantages of DEAI and its superiority to centralized solutions is a direct, in-depth comparison between centralized and decentralized solutions that work toward the same goal or can be used interchangeably \cite{somy2019ownership}. For example, a decentralized approach can be 10\% slower, but the users have better privacy and transparency of the AI system. Nevertheless, the decentralized solutions need additional evaluation metrics, most often related to security, such as the probability of different attacks, immunity to malicious participants, and other common attacks in decentralized networks \cite{teerapittayanon2019daimon, boubouh2020robust}.

\section{Discussion}
\label{sec:discussion}

As evident in the previous sections, there is a lot of existing work on decentralizing AI, although often scattered to a certain degree. After identifying all the building blocks, with some being researched as a standalone component and some already part of DEAI networks and systems, we can formulate the definition of DEAI network (RQ 1): \textit{DEAI network is a decentralized system that provides an environment for AI agents to operate and communicate in it, offer endpoints that end-users or other AI agents can call}. The high-level definition was formulated based on our comprehensive research of various proposed architectures and, to some degree, implemented projects in the gray literature. For the definition and the general concepts to be fully validated, the proposed network must be deployed and started (often referred to as the network's genesis). Through multiple years of running in the production and being used permissionless by different actors, the solutions can be fully validated and performance measured. None of the analyzed worked proposed architecture that contained all the building blocks or most of them while still being fully validated in the production (RQ 2). Nevertheless, some works, such as \cite{singularitynet, effect_network, kumar2020marketplace, pds2}, proposed fully-fledged architectures that (at least in theory) encapsulate the ideas from many identified building blocks in this paper.

During the in-depth review and analysis, we identified multiple advantages and disadvantages when comparing DEAI solutions to CEAI (RQ 3). The most significant benefits are censorship resistance, no single point of failure, increased interoperability and communication possibilities between different solutions, predictability, and enhanced democratized access to technologies. Another improvement to the AI field is increased transparency in several steps of the AI workflows, which are becoming increasingly blurry in the CEAI due to their business models. But at the cost of the advantages, disadvantages can also be identified, such as slower development time and decision-making process, lower scalability and performance of current solutions today, and earlier stage of solutions. One of CEAI's advantages is that it is farther ahead in the development and adoption cycle; thus, it is hard to perform a fair comparison, though the interest in DEAI is growing each year.

\begin{figure}[!ht]
\centering
\includegraphics[width=\columnwidth]{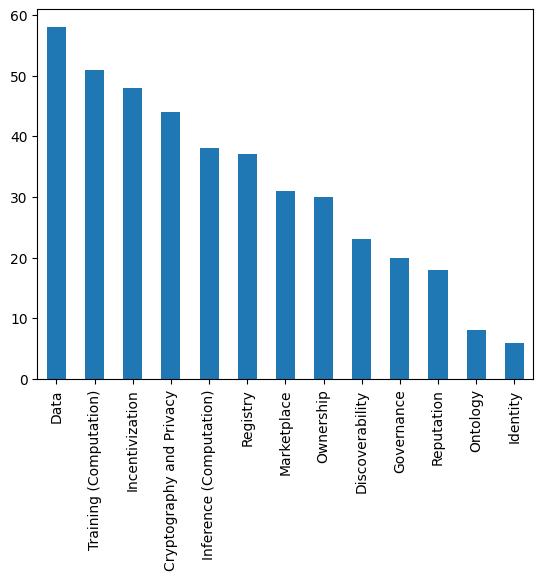}
\caption{The number of papers per each building block.}
\label{fig:building_blocks}
\end{figure}

This paper identified 13 essential building blocks contributing important features or characteristics that benefit the DEAI network (RQ 4). While some of the building blocks are explored in the literature or are significantly related to other fields, such as data and marketplaces, several are still unexplored, and researchers should focus more on them, such as \textbf{\textit{identity}}, \textbf{\textit{ontology}}, and \textbf{\textit{reputation}}. More work should also be focused on evaluation methods and proposing standardized approaches for evaluating the performance of different solutions, improving fair comparisons, and defining the optimization goals. Figure \ref{fig:building_blocks} shows how many papers we identified for each building block. We also identified three features and one challenge besides the building blocks. The analyzed features were not marked as building blocks since they are not separate components of the solution but rather the guidelines or features that should be considered when designing each building block. The identified challenge - updating global models - is one of the challenges for several building blocks, such as registry, training (computation), and incentivization.

\begin{figure}[!ht]
\centering
\includegraphics[width=\columnwidth]{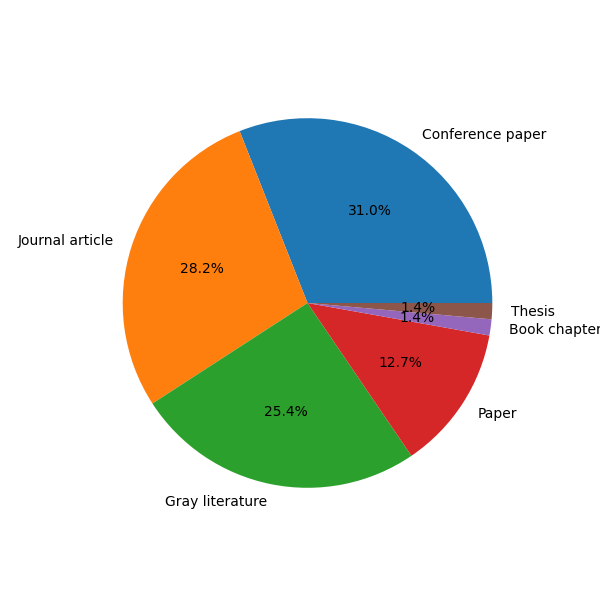}
\caption{Percentages of papers based on place of publication.}
\label{fig:plac_of_publication}
\end{figure}

\begin{figure}[!ht]
\centering
\includegraphics[width=\columnwidth]{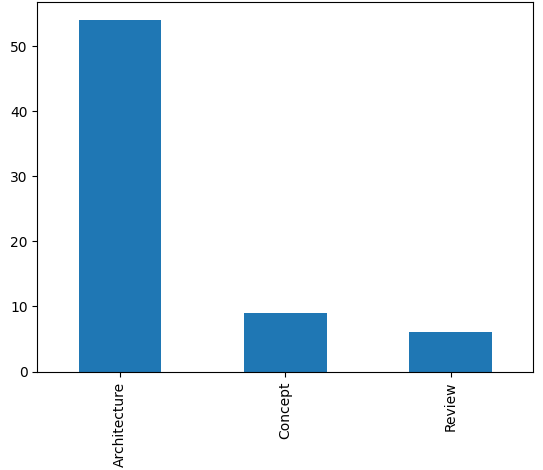}
\caption{The number of papers based on contribution.}
\label{fig:contribution}
\end{figure}

While the main emphasis of the paper was the identification of building blocks as well as features, challenges, and evaluation methods for solutions, we also screened additional attributes for further analysis. These attributes include publication place, contribution, database, year, supported AI models, and system/network type. Most of the identified work came from journal articles, gray literature, and conference papers (Figure~\ref{fig:plac_of_publication}). Similar to other web and decentralization technologies fields, much progress and state-of-the-art are often presented in the gray literature since the validation and effectiveness of the systems must often be validated on actual wide-scale deployments. Out of the identified articles, most of them were focused on defining the technical architecture or proposing new concepts (Figure~\ref{fig:contribution}).

\begin{figure}[!ht]
\centering
\includegraphics[width=\columnwidth]{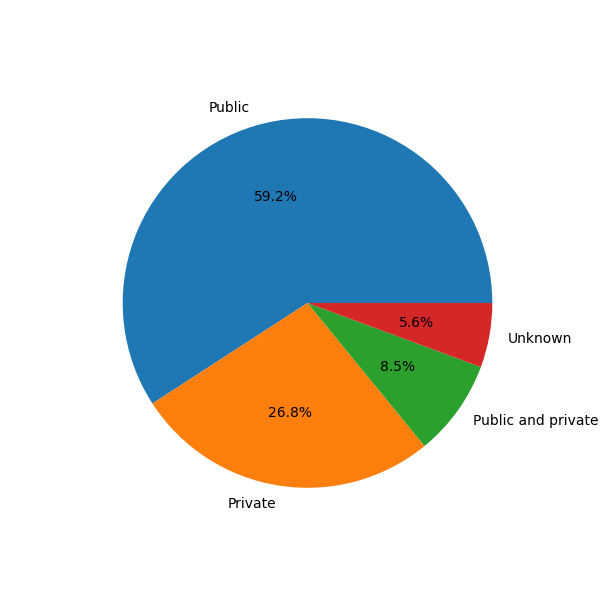}
\caption{Percentages of papers based on network type.}
\label{fig:network_type}
\end{figure}

While a lot of works tried to abstract the systems to support a wide range of AI models, some solutions were explicitly designed for ML and DL models (e.g., training and inference - computation), which can be attributed to the popularity of these models and the fact that most AI systems in producing rely only on these type of models. Another screened attribute was whether the system is public or private (or permissioned) regarding node operators or implementation of smart contracts, meaning anyone can join the network without any obligations (Figure~\ref{fig:network_type}). While the P2P and other decentralized networks evolved and were created due to being public and without a controlling entity, several works proposed solutions based on private networks/blockchains, and the ideas still contribute to the broader field of DEAI.

\begin{figure}[!ht]
\centering
\includegraphics[width=\columnwidth]{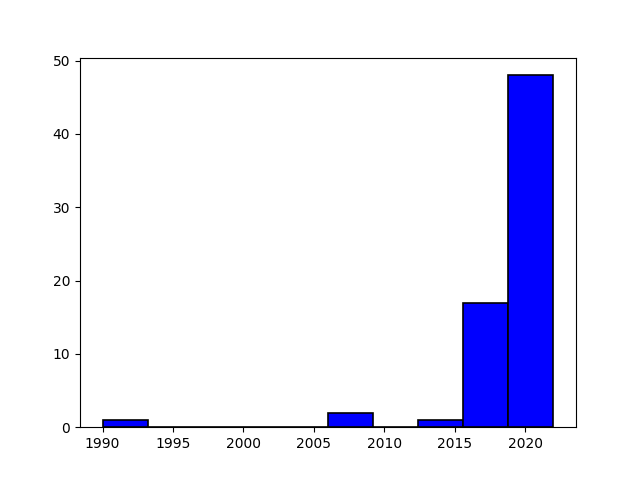}
\caption{The number of papers per year.}
\label{fig:year}
\end{figure}

As seen in Figure~\ref{fig:year}, most identified papers are from the last few years, more precisely from 2017. This can be attributed to the broader recognition of blockchain technology that year. While the first paper that mentioned the term DEAI is from 1990, the term first referred to MAS, as well as some articles in the 2000s that focus on the implementation and designs of MAS systems. The field is getting higher recognition because of the significant advancements in P2P systems and other decentralized technologies, such as blockchain and decentralized storage networks.

One of the main goals of the paper was to showcase how the development of DEAI networks and solutions can be approached from a different perspective - bottom-up instead of top-down. This means that instead of focusing on the whole solution, there should be research on only one building block at a time, exploring how this single building block fits together with other building blocks. This type of research and development is common in other areas, such as decentralized identity, where a lot of effort is put into defining standards and low-level building blocks and how they fit together and can be used in high-level solutions and systems. There should also be more cooperation between academics and companies from the industry working on the DEAI projects since AI is very popular and researched in the academic sphere. At the same time, the DEAI is predominated in the industry. While some solutions can be defined and tested in the articles and laboratory environment at the beginning, to achieve production readiness and prove to be effective, the P2P, decentralized, and permissionless solutions must be tested and tried in the real world scenarios.

\section{Conclusion}
\label{sec:conclusion}

This paper reviewed the latest research and development advances in the DEAI field through the SLR research methodology. We identified 71 articles and projects that were relevant to our analysis. The purpose of SLR was to analyze the latest approaches to DEAI networks and systems, identify evaluation methods for these kinds of systems, and identify core building blocks that constitute these solutions.

The study revealed there is comprehensive literature available on the DEAI, both in the form of scientific papers and gray literature. We identified relevant work for each of the 13 building blocks, although some are being explored further in the development phase while some are still in the early stages. We also identified features, challenges, and evaluation methods alongside the identified building blocks. There is still a lot of improvement in the research domain, where more real-world experiments, implementations, and validation are needed to validate solutions presented in scientific papers. On the other hand, projects from gray literature should explore defining the components and building blocks for DEAI networks from the bottom-up, focusing on core building blocks and how they can be used together to construct high-level abstractions and whole systems, enhancing cooperation and interoperability between different systems.

\section*{Acknowledgments}
\label{sec:ack}

This work was supported by the Slovenian Research Agency (Research Core Funding) under Grant P2-0057.

\bibliographystyle{IEEEtran}
\bibliography{main}

\begin{thebibliography}{100}
\providecommand{\url}[1]{#1}
\csname url@samestyle\endcsname
\providecommand{\newblock}{\relax}
\providecommand{\bibinfo}[2]{#2}
\providecommand{\BIBentrySTDinterwordspacing}{\spaceskip=0pt\relax}
\providecommand{\BIBentryALTinterwordstretchfactor}{4}
\providecommand{\BIBentryALTinterwordspacing}{\spaceskip=\fontdimen2\font plus
\BIBentryALTinterwordstretchfactor\fontdimen3\font minus
  \fontdimen4\font\relax}
\providecommand{\BIBforeignlanguage}[2]{{%
\expandafter\ifx\csname l@#1\endcsname\relax
\typeout{** WARNING: IEEEtran.bst: No hyphenation pattern has been}%
\typeout{** loaded for the language `#1'. Using the pattern for}%
\typeout{** the default language instead.}%
\else
\language=\csname l@#1\endcsname
\fi
#2}}
\providecommand{\BIBdecl}{\relax}
\BIBdecl

\bibitem{vaswani2017attention}
A.~Vaswani, N.~Shazeer, N.~Parmar, J.~Uszkoreit, L.~Jones, A.~N. Gomez,
  {\L}.~Kaiser, and I.~Polosukhin, ``Attention is all you need,''
  \emph{Advances in neural information processing systems}, vol.~30, 2017.

\bibitem{jordan2015machine}
M.~I. Jordan and T.~M. Mitchell, ``Machine learning: Trends, perspectives, and
  prospects,'' \emph{Science}, vol. 349, no. 6245, pp. 255--260, 2015.

\bibitem{moura2020clouding}
G.~C. Moura, S.~Castro, W.~Hardaker, M.~Wullink, and C.~Hesselman, ``Clouding
  up the internet: How centralized is dns traffic becoming?'' in
  \emph{Proceedings of the ACM Internet Measurement Conference}, 2020, pp.
  42--49.

\bibitem{nakamoto2008bitcoin}
S.~Nakamoto, ``Bitcoin whitepaper,'' \emph{URL: https://bitcoin. org/bitcoin.
  pdf-(: 17.07. 2019)}, 2008.

\bibitem{buterin2014next}
V.~Buterin \emph{et~al.}, ``A next-generation smart contract and decentralized
  application platform,'' \emph{white paper}, vol.~3, no.~37, pp. 2--1, 2014.

\bibitem{ipfs}
\BIBentryALTinterwordspacing
``Ipfs docs.'' [Online]. Available: \url{https://docs.ipfs.tech/}
\BIBentrySTDinterwordspacing

\bibitem{filecoin}
\BIBentryALTinterwordspacing
``Filecoin docs.'' [Online]. Available: \url{https://docs.filecoin.io/}
\BIBentrySTDinterwordspacing

\bibitem{miiller1990decentralized}
Y.~Miiller, ``Decentralized artificial intelligence,'' \emph{Decentralised AI},
  pp. 3--13, 1990.

\bibitem{chaib1992trends}
B.~Chaib-Draa, B.~Moulin, R.~Mandiau, and P.~Millot, ``Trends in distributed
  artificial intelligence,'' \emph{Artificial Intelligence Review}, vol.~6, pp.
  35--66, 1992.

\bibitem{tagde2021blockchain}
P.~Tagde, S.~Tagde, T.~Bhattacharya, P.~Tagde, H.~Chopra, R.~Akter, D.~Kaushik,
  and M.~H. Rahman, ``Blockchain and artificial intelligence technology in
  e-health,'' \emph{Environmental Science and Pollution Research}, vol.~28, pp.
  52\,810--52\,831, 2021.

\bibitem{shafay2023blockchain}
M.~Shafay, R.~W. Ahmad, K.~Salah, I.~Yaqoob, R.~Jayaraman, and M.~Omar,
  ``Blockchain for deep learning: review and open challenges,'' \emph{Cluster
  Computing}, vol.~26, no.~1, pp. 197--221, 2023.

\bibitem{ai_and_blockchain_integration_ieee}
B.~Chavali, S.~K. Khatri, and S.~A. Hossain, ``Ai and blockchain integration,''
  in \emph{2020 8th International Conference on Reliability, Infocom
  Technologies and Optimization (Trends and Future Directions) (ICRITO)}, 2020,
  pp. 548--552.

\bibitem{survey_ai_ml}
Y.~Liu, F.~R. Yu, X.~Li, H.~Ji, and V.~C.~M. Leung, ``Blockchain and machine
  learning for communications and networking systems,'' \emph{IEEE
  Communications Surveys \& Tutorials}, vol.~22, no.~2, pp. 1392--1431, 2020.

\bibitem{blockchain_ml_iot}
A.~Miglani and N.~Kumar, ``Blockchain management and machine learning
  adaptation for iot environment in 5g and beyond networks: A systematic
  review,'' \emph{Computer Communications}, vol. 178, pp. 37--63, 2021.

\bibitem{wang2021applications}
R.~Wang, M.~Luo, Y.~Wen, L.~Wang, K.-K. Raymond~Choo, and D.~He, ``The
  applications of blockchain in artificial intelligence,'' \emph{Security and
  Communication Networks}, vol. 2021, pp. 1--16, 2021.

\bibitem{hashim2019decrypting}
M.~Hashim, ``Decrypting practical ai: empowering or enslaving humans?''
  \emph{Accessed: Mar}, vol.~24, p. 2021, 2019.

\bibitem{daniel2022ipfs}
E.~Daniel and F.~Tschorsch, ``Ipfs and friends: A qualitative comparison of
  next generation peer-to-peer data networks,'' \emph{IEEE Communications
  Surveys \& Tutorials}, vol.~24, no.~1, pp. 31--52, 2022.

\bibitem{russell2010artificial}
S.~J. Russell, \emph{Artificial intelligence a modern approach}.\hskip 1em plus
  0.5em minus 0.4em\relax Pearson Education, Inc., 2010.

\bibitem{krizhevsky2012imagenet}
A.~Krizhevsky, I.~Sutskever, and G.~E. Hinton, ``Imagenet classification with
  deep convolutional neural networks,'' \emph{Advances in neural information
  processing systems}, vol.~25, 2012.

\bibitem{kipf2016semi}
T.~N. Kipf and M.~Welling, ``Semi-supervised classification with graph
  convolutional networks,'' \emph{arXiv preprint arXiv:1609.02907}, 2016.

\bibitem{graves2012long}
A.~Graves and A.~Graves, ``Long short-term memory,'' \emph{Supervised sequence
  labelling with recurrent neural networks}, pp. 37--45, 2012.

\bibitem{montes2019distributed}
G.~A. Montes and B.~Goertzel, ``Distributed, decentralized, and democratized
  artificial intelligence,'' \emph{Technological Forecasting and Social
  Change}, vol. 141, pp. 354--358, 2019.

\bibitem{stanford_index}
\BIBentryALTinterwordspacing
``Ai index report 2023 – artificial intelligence index.'' [Online].
  Available: \url{https://aiindex.stanford.edu/report/}
\BIBentrySTDinterwordspacing

\bibitem{o1996foundations}
G.~M. O'Hare and N.~R. Jennings, \emph{Foundations of distributed artificial
  intelligence}.\hskip 1em plus 0.5em minus 0.4em\relax John Wiley \& Sons,
  1996, vol.~9.

\bibitem{gupta2018distributed}
O.~Gupta and R.~Raskar, ``Distributed learning of deep neural network over
  multiple agents,'' \emph{Journal of Network and Computer Applications}, vol.
  116, pp. 1--8, 2018.

\bibitem{chahal2020hitchhiker}
K.~S. Chahal, M.~S. Grover, K.~Dey, and R.~R. Shah, ``A hitchhiker’s guide on
  distributed training of deep neural networks,'' \emph{Journal of Parallel and
  Distributed Computing}, vol. 137, pp. 65--76, 2020.

\bibitem{mas_survey}
A.~Dorri, S.~S. Kanhere, and R.~Jurdak, ``Multi-agent systems: A survey,''
  \emph{IEEE Access}, vol.~6, pp. 28\,573--28\,593, 2018.

\bibitem{jiang2013understanding}
Y.~Jiang and J.~Jiang, ``Understanding social networks from a multiagent
  perspective,'' \emph{IEEE Transactions on Parallel and Distributed Systems},
  vol.~25, no.~10, pp. 2743--2759, 2013.

\bibitem{nguyen2012agent}
C.~P. Nguyen and A.~J. Flueck, ``Agent based restoration with distributed
  energy storage support in smart grids,'' \emph{IEEE Transactions on Smart
  Grid}, vol.~3, no.~2, pp. 1029--1038, 2012.

\bibitem{zhou2019edge}
Z.~Zhou, X.~Chen, E.~Li, L.~Zeng, K.~Luo, and J.~Zhang, ``Edge intelligence:
  Paving the last mile of artificial intelligence with edge computing,''
  \emph{Proceedings of the IEEE}, vol. 107, no.~8, pp. 1738--1762, 2019.

\bibitem{federated_learning}
T.~Li, A.~K. Sahu, A.~Talwalkar, and V.~Smith, ``Federated learning:
  Challenges, methods, and future directions,'' \emph{IEEE Signal Processing
  Magazine}, vol.~37, no.~3, pp. 50--60, 2020.

\bibitem{rauchs2018distributed}
M.~Rauchs, A.~Glidden, B.~Gordon, G.~C. Pieters, M.~Recanatini, F.~Rostand,
  K.~Vagneur, and B.~Z. Zhang, ``Distributed ledger technology systems: A
  conceptual framework,'' \emph{Available at SSRN 3230013}, 2018.

\bibitem{belk2022money}
R.~Belk, M.~Humayun, and M.~Brouard, ``Money, possessions, and ownership in the
  metaverse: Nfts, cryptocurrencies, web3 and wild markets,'' \emph{Journal of
  Business Research}, vol. 153, pp. 198--205, 2022.

\bibitem{thegraph}
\BIBentryALTinterwordspacing
``Get started - the graph docs.'' [Online]. Available:
  \url{https://thegraph.com/docs/en/}
\BIBentrySTDinterwordspacing

\bibitem{deai_web3}
L.~Cao, ``Decentralized ai: Edge intelligence and smart blockchain, metaverse,
  web3, and desci,'' \emph{IEEE Intelligent Systems}, vol.~37, no.~3, pp.
  6--19, 2022.

\bibitem{yu2022blockchain}
F.~Yu, H.~Lin, X.~Wang, A.~Yassine, and M.~S. Hossain, ``Blockchain-empowered
  secure federated learning system: Architecture and applications,''
  \emph{Computer Communications}, vol. 196, pp. 55--65, 2022.

\bibitem{dorri2018multi}
A.~Dorri, S.~S. Kanhere, and R.~Jurdak, ``Multi-agent systems: A survey,''
  \emph{Ieee Access}, vol.~6, pp. 28\,573--28\,593, 2018.

\bibitem{ponomarev2017multi}
S.~Ponomarev and A.~Voronkov, ``Multi-agent systems and decentralized
  artificial superintelligence,'' \emph{arXiv preprint arXiv:1702.08529}, 2017.

\bibitem{Kitchenham2009}
B.~Kitchenham, O.~P. Brereton, D.~Budgen, M.~Turner, J.~Bailey, and S.~Linkman,
  ``Systematic literature reviews in software engineering – a systematic
  literature review,'' \emph{Information and Software Technology}, vol.~51, pp.
  7--15, 2009.

\bibitem{hugging_face_spaces}
\BIBentryALTinterwordspacing
``Spaces - hugging face.'' [Online]. Available:
  \url{https://huggingface.co/spaces}
\BIBentrySTDinterwordspacing

\bibitem{singularitynet}
\BIBentryALTinterwordspacing
``Welcome to the ai dev community, powered by open collaboration.'' [Online].
  Available: \url{https://dev.singularitynet.io/}
\BIBentrySTDinterwordspacing

\bibitem{dinemmo}
A.~Marathe, K.~Narayanan, A.~Gupta, and M.~P.R., ``Dinemmo: Decentralized
  incentivization for enterprise marketplace models,'' in \emph{2018 IEEE 25th
  International Conference on High Performance Computing Workshops (HiPCW)},
  2018, pp. 95--100.

\bibitem{fabricfl}
V.~Mothukuri, R.~M. Parizi, S.~Pouriyeh, A.~Dehghantanha, and K.-K.~R. Choo,
  ``Fabricfl: Blockchain-in-the-loop federated learning for trusted
  decentralized systems,'' \emph{IEEE Systems Journal}, vol.~16, no.~3, pp.
  3711--3722, 2022.

\bibitem{trustless_api}
V.~Arya, S.~Sen, and P.~Kodeswaran, ``Blockchain enabled trustless api
  marketplace,'' in \emph{2020 International Conference on COMmunication
  Systems \& NETworkS (COMSNETS)}, 2020, pp. 731--735.

\bibitem{masurkar2019decentralized}
A.~S. Masurkar, ``Decentralized artificial intelligence with data privacy
  protection,'' 2019.

\bibitem{hackathon_ai}
\BIBentryALTinterwordspacing
``Decentralized ai and data marketplace | devfolio.'' [Online]. Available:
  \url{https://devfolio.co/projects/decentralized-ai-and-data-marketplace-fa56}
\BIBentrySTDinterwordspacing

\bibitem{adel2022decentralizing}
K.~Adel, A.~Elhakeem, and M.~Marzouk, ``Decentralizing construction ai
  applications using blockchain technology,'' \emph{Expert Systems with
  Applications}, vol. 194, p. 116548, 2022.

\bibitem{ocean_protocol}
\BIBentryALTinterwordspacing
``Orientation - ocean protocol.'' [Online]. Available:
  \url{https://docs.oceanprotocol.com/}
\BIBentrySTDinterwordspacing

\bibitem{proof_of_learning}
F.~Bravo-Marquez, S.~Reeves, and M.~Ugarte, ``Proof-of-learning: A blockchain
  consensus mechanism based on machine learning competitions,'' in \emph{2019
  IEEE International Conference on Decentralized Applications and
  Infrastructures (DAPPCON)}, 2019, pp. 119--124.

\bibitem{fetch.ai}
\BIBentryALTinterwordspacing
``At a glance - developer documentation.'' [Online]. Available:
  \url{https://docs.fetch.ai/}
\BIBentrySTDinterwordspacing

\bibitem{blythman2022libraries}
R.~Blythman, M.~Arshath, J.~Sm{\'e}kal, H.~Shaji, S.~Vivona, and T.~Dunmore,
  ``Libraries, integrations and hubs for decentralized ai using ipfs,''
  \emph{arXiv preprint arXiv:2210.16651}, 2022.

\bibitem{wang2021non}
Q.~Wang, R.~Li, Q.~Wang, and S.~Chen, ``Non-fungible token (nft): Overview,
  evaluation, opportunities and challenges,'' \emph{arXiv preprint
  arXiv:2105.07447}, 2021.

\bibitem{deepbrainchain}
\BIBentryALTinterwordspacing
``Deepbrainchain whitepaper.'' [Online]. Available:
  \url{https://www.deepbrainchain.org/assets/pdf/DeepBrainChainWhitepaper_en.pdf}
\BIBentrySTDinterwordspacing

\bibitem{oraichain}
\BIBentryALTinterwordspacing
``Whitepaper - oraichain.'' [Online]. Available: \url{https://docs.orai.io/}
\BIBentrySTDinterwordspacing

\bibitem{platon}
\BIBentryALTinterwordspacing
``Overview | platon.'' [Online]. Available:
  \url{https://devdocs.platon.network/docs/en/}
\BIBentrySTDinterwordspacing

\bibitem{giza}
\BIBentryALTinterwordspacing
``Orion - orion.'' [Online]. Available:
  \url{https://orion.gizatech.xyz/welcome/readme}
\BIBentrySTDinterwordspacing

\bibitem{cortex}
\BIBentryALTinterwordspacing
``Cortex overview.'' [Online]. Available:
  \url{https://github.com/CortexFoundation/tech-doc/blob/master/cortex-details.md}
\BIBentrySTDinterwordspacing

\bibitem{sketching}
A.~Kumar, B.~Finley, T.~Braud, S.~Tarkoma, and P.~Hui, ``Sketching an ai
  marketplace: Tech, economic, and regulatory aspects,'' \emph{IEEE Access},
  vol.~9, pp. 13\,761--13\,774, 2021.

\bibitem{lu2018enabling}
Y.~Lu, Q.~Tang, and G.~Wang, ``On enabling machine learning tasks atop public
  blockchains: A crowdsourcing approach,'' in \emph{2018 IEEE international
  conference on data mining workshops (ICDMW)}.\hskip 1em plus 0.5em minus
  0.4em\relax IEEE, 2018, pp. 81--88.

\bibitem{algovera_ai}
\BIBentryALTinterwordspacing
``Docs - algovera ai.'' [Online]. Available: \url{https://docs.algovera.ai/}
\BIBentrySTDinterwordspacing

\bibitem{micro_coll}
J.~D. Harris and B.~Waggoner, ``Decentralized and collaborative ai on
  blockchain,'' in \emph{2019 IEEE International Conference on Blockchain
  (Blockchain)}, 2019, pp. 368--375.

\bibitem{nguyen2021marketplace}
L.~D. Nguyen, S.~R. Pandey, S.~Beatriz, A.~Broering, and P.~Popovski, ``A
  marketplace for trading ai models based on blockchain and incentives for iot
  data,'' \emph{arXiv preprint arXiv:2112.02870}, 2021.

\bibitem{johnson2021ichain}
C.~Johnson, T.~Lu, P.~Rivera, D.~McDonald, S.~Pritchett, and L.~Peng, ``ichain:
  Peer-to-peer machine learning powered by blockchain technology,''
  \emph{Frontiers in Blockchain}, vol.~4, p. 676159, 2021.

\bibitem{learning_markets}
L.~Ouyang, Y.~Yuan, and F.-Y. Wang, ``Learning markets: An ai collaboration
  framework based on blockchain and smart contracts,'' \emph{IEEE Internet of
  Things Journal}, vol.~9, no.~16, pp. 14\,273--14\,286, 2022.

\bibitem{liang2021omnilytics}
J.~Liang, S.~Li, B.~Cao, W.~Jiang, and C.~He, ``Omnilytics: A blockchain-based
  secure data market for decentralized machine learning,'' \emph{arXiv preprint
  arXiv:2107.05252}, 2021.

\bibitem{numerai}
\BIBentryALTinterwordspacing
``Overview - numerai tournament.'' [Online]. Available:
  \url{https://docs.numer.ai/numerai-tournament/readme}
\BIBentrySTDinterwordspacing

\bibitem{kurtulmus2018trustless}
A.~B. Kurtulmus and K.~Daniel, ``Trustless machine learning contracts;
  evaluating and exchanging machine learning models on the ethereum
  blockchain,'' \emph{arXiv preprint arXiv:1802.10185}, 2018.

\bibitem{weng2019deepchain}
J.~Weng, J.~Weng, J.~Zhang, M.~Li, Y.~Zhang, and W.~Luo, ``Deepchain: Auditable
  and privacy-preserving deep learning with blockchain-based incentive,''
  \emph{IEEE Transactions on Dependable and Secure Computing}, vol.~18, no.~5,
  pp. 2438--2455, 2019.

\bibitem{pds2}
L.~Giaretta, I.~Savvidis, T.~Marchioro, S.~Girdzijauskas, G.~Pallis, M.~D.
  Dikaiakos, and E.~Markatos, ``Pds2: A user-centered decentralized marketplace
  for privacy preserving data processing,'' in \emph{2021 IEEE 37th
  International Conference on Data Engineering Workshops (ICDEW)}, 2021, pp.
  92--99.

\bibitem{kim2019efficient}
H.~Kim, S.-H. Kim, J.~Y. Hwang, and C.~Seo, ``Efficient privacy-preserving
  machine learning for blockchain network,'' \emph{Ieee Access}, vol.~7, pp.
  136\,481--136\,495, 2019.

\bibitem{rao2020bittensor}
Y.~Rao, J.~Steeves, A.~Shaabana, D.~Attevelt, and M.~McAteer, ``Bittensor: A
  peer-to-peer intelligence market,'' \emph{arXiv preprint arXiv:2003.03917},
  2020.

\bibitem{boubouh2020robust}
K.~Boubouh, A.~Boussetta, Y.~Benkaouz, and R.~Guerraoui, ``Robust p2p
  personalized learning,'' in \emph{2020 International Symposium on Reliable
  Distributed Systems (SRDS)}.\hskip 1em plus 0.5em minus 0.4em\relax IEEE,
  2020, pp. 299--308.

\bibitem{kusi2020training}
G.~A. Kusi, Q.~Xia, C.~N.~A. Cobblah, J.~Gao, and H.~Xia, ``Training machine
  learning models through preserved decentralization,'' in \emph{2020 16th
  International Conference on Mobility, Sensing and Networking (MSN)}.\hskip
  1em plus 0.5em minus 0.4em\relax IEEE, 2020, pp. 465--472.

\bibitem{teerapittayanon2019daimon}
S.~Teerapittayanon and H.~Kung, ``Daimon: A decentralized artificial
  intelligence model network,'' in \emph{2019 IEEE International Conference on
  Blockchain (Blockchain)}.\hskip 1em plus 0.5em minus 0.4em\relax IEEE, 2019,
  pp. 132--139.

\bibitem{li2020blockchain}
M.~Li, Q.~Wang, and W.~Zhang, ``Blockchain-based distributed machine learning
  towards statistical challenges,'' in \emph{Blockchain and Trustworthy
  Systems: Second International Conference, BlockSys 2020, Dali, China, August
  6--7, 2020, Revised Selected Papers 2}.\hskip 1em plus 0.5em minus
  0.4em\relax Springer, 2020, pp. 549--564.

\bibitem{effect_network}
\BIBentryALTinterwordspacing
``Effect network white paper.'' [Online]. Available:
  \url{https://effect.network/download/effect_whitepaper.pdf}
\BIBentrySTDinterwordspacing

\bibitem{somy2019ownership}
N.~B. Somy, K.~Kannan, V.~Arya, S.~Hans, A.~Singh, P.~Lohia, and S.~Mehta,
  ``Ownership preserving ai market places using blockchain,'' in \emph{2019
  IEEE international conference on blockchain (Blockchain)}.\hskip 1em plus
  0.5em minus 0.4em\relax IEEE, 2019, pp. 156--165.

\bibitem{nunet}
\BIBentryALTinterwordspacing
``Nunet platform.'' [Online]. Available:
  \url{https://docs.nunet.io/infohub/state-of-nunet/nunet-platform}
\BIBentrySTDinterwordspacing

\bibitem{douceur2002sybil}
J.~R. Douceur, ``The sybil attack,'' in \emph{International workshop on
  peer-to-peer systems}.\hskip 1em plus 0.5em minus 0.4em\relax Springer, 2002,
  pp. 251--260.

\bibitem{blanchard2017machine}
P.~Blanchard, E.~M. El~Mhamdi, R.~Guerraoui, and J.~Stainer, ``Machine learning
  with adversaries: Byzantine tolerant gradient descent,'' \emph{Advances in
  neural information processing systems}, vol.~30, 2017.

\bibitem{peyvandi2022privacy}
A.~Peyvandi, B.~Majidi, S.~Peyvandi, and J.~C. Patra, ``Privacy-preserving
  federated learning for scalable and high data quality
  computational-intelligence-as-a-service in society 5.0,'' \emph{Multimedia
  Tools and Applications}, vol.~81, no.~18, pp. 25\,029--25\,050, 2022.

\bibitem{vickery1997ontologies}
B.~C. Vickery, ``Ontologies,'' \emph{Journal of information science}, vol.~23,
  no.~4, pp. 277--286, 1997.

\bibitem{tran2008mobmas}
Q.-N.~N. Tran and G.~Low, ``Mobmas: A methodology for ontology-based
  multi-agent systems development,'' \emph{Information and Software
  Technology}, vol.~50, no. 7-8, pp. 697--722, 2008.

\bibitem{tran2008preliminary}
Q.-N.~N. Tran, G.~Beydoun, G.~Low, and C.~Gonzalez-Perez, ``Preliminary
  validation of mobmas (ontology-centric agent oriented methodology): design of
  a peer-to-peer information sharing mas,'' in \emph{Agent-Oriented Information
  Systems IV: 8th International Bi-Conference Workshop, AOIS 2006, Hakodate,
  Japan, May 9, 2006 and Luxembourg, Luxembourg, June 6, 2006, Revised Selected
  Papers}.\hskip 1em plus 0.5em minus 0.4em\relax Springer, 2008, pp. 73--89.

\bibitem{tran2007design}
N.~Tran, G.~Beydoun, and G.~Low, ``Design of a peer-to-peer information sharing
  mas using mobmas (ontology-centric agent oriented methodology),'' in
  \emph{Advances in information systems development: New methods and practice
  for the networked society}.\hskip 1em plus 0.5em minus 0.4em\relax Springer,
  2007, pp. 63--76.

\bibitem{gorodetskii2008development}
V.~Gorodetskii, O.~Karsaev, V.~Samoilov, and S.~Serebryakov, ``Development
  tools for open agent networks,'' \emph{Journal of Computer and Systems
  Sciences International}, vol.~47, pp. 429--446, 2008.

\bibitem{bacalhau}
\BIBentryALTinterwordspacing
``Bacalhau computer over data.'' [Online]. Available:
  \url{https://www.bacalhau.org/}
\BIBentrySTDinterwordspacing

\bibitem{goel2019deepring}
A.~Goel, A.~Agarwal, M.~Vatsa, R.~Singh, and N.~Ratha, ``Deepring: Protecting
  deep neural network with blockchain,'' in \emph{Proceedings of the IEEE/CVF
  conference on computer vision and pattern recognition workshops}, 2019, pp.
  0--0.

\bibitem{kuo2018modelchain}
T.-T. Kuo and L.~Ohno-Machado, ``Modelchain: Decentralized privacy-preserving
  healthcare predictive modeling framework on private blockchain networks,''
  \emph{arXiv preprint arXiv:1802.01746}, 2018.

\bibitem{du2022accelerating}
Y.~Du, C.~Leung, Z.~Wang, and V.~C. Leung, ``Accelerating blockchain-enabled
  distributed machine learning by proof of useful work,'' in \emph{2022
  IEEE/ACM 30th International Symposium on Quality of Service (IWQoS)}.\hskip
  1em plus 0.5em minus 0.4em\relax IEEE, 2022, pp. 1--10.

\bibitem{colink}
\BIBentryALTinterwordspacing
``Welcome to colink.'' [Online]. Available:
  \url{https://co-learn.notion.site/co-learn/Welcome-to-CoLink-5bf0c431201441e68cba07a5f7101728}
\BIBentrySTDinterwordspacing

\bibitem{zhou2020pirate}
S.~Zhou, H.~Huang, W.~Chen, P.~Zhou, Z.~Zheng, and S.~Guo, ``Pirate: A
  blockchain-based secure framework of distributed machine learning in 5g
  networks,'' \emph{IEEE Network}, vol.~34, no.~6, pp. 84--91, 2020.

\bibitem{zhang2020sablockfl}
Z.~Zhang, T.~Yang, and Y.~Liu, ``Sablockfl: a blockchain-based smart agent
  system architecture and its application in federated learning,''
  \emph{International Journal of Crowd Science}, vol.~4, no.~2, pp. 133--147,
  2020.

\bibitem{majeed2019flchain}
U.~Majeed and C.~S. Hong, ``Flchain: Federated learning via mec-enabled
  blockchain network,'' in \emph{2019 20th Asia-Pacific Network Operations and
  Management Symposium (APNOMS)}.\hskip 1em plus 0.5em minus 0.4em\relax IEEE,
  2019, pp. 1--4.

\bibitem{clifton2022decentralized}
C.~Clifton, R.~Blythman, and K.~Tulusan, ``Is decentralized ai safer?'' 2022.

\bibitem{kumar2020marketplace}
A.~Kumar, B.~Finley, T.~Braud, S.~Tarkoma, and P.~Hui, ``Marketplace for ai
  models,'' \emph{arXiv preprint arXiv:2003.01593}, 2020.

\bibitem{ai_bl_integration}
B.~Chavali, S.~K. Khatri, and S.~A. Hossain, ``Ai and blockchain integration,''
  in \emph{2020 8th International Conference on Reliability, Infocom
  Technologies and Optimization (Trends and Future Directions) (ICRITO)}, 2020,
  pp. 548--552.

\bibitem{gentry2009fully}
C.~Gentry, \emph{A fully homomorphic encryption scheme}.\hskip 1em plus 0.5em
  minus 0.4em\relax Stanford university, 2009.

\bibitem{privacy_svm}
M.~Shen, X.~Tang, L.~Zhu, X.~Du, and M.~Guizani, ``Privacy-preserving support
  vector machine training over blockchain-based encrypted iot data in smart
  cities,'' \emph{IEEE Internet of Things Journal}, vol.~6, no.~5, pp.
  7702--7712, 2019.

\bibitem{trusted_collaborations}
K.~Kannan, A.~Singh, M.~Verma, P.~Jayachandran, and S.~Mehta,
  ``Blockchain-based platform for trusted collaborations on data and ai
  models,'' in \emph{2020 IEEE International Conference on Blockchain
  (Blockchain)}, 2020, pp. 82--89.

\bibitem{pelt2021defining}
R.~v. Pelt, S.~Jansen, D.~Baars, and S.~Overbeek, ``Defining blockchain
  governance: A framework for analysis and comparison,'' \emph{Information
  Systems Management}, vol.~38, no.~1, pp. 21--41, 2021.

\bibitem{kervsivc2022using}
V.~Ker{\v{s}}i{\v{c}}, A.~Vre{\v{c}}ko, U.~Vidovi{\v{c}}, M.~Domajnko, and
  M.~Turkanovi{\'c}, ``Using self-sovereign-identity principles to prove your
  worth in decentralized autonomous organizations,'' \emph{Proceedings
  http://ceur-ws. org ISSN}, vol. 1613, p. 0073, 2022.

\bibitem{trent_ai_daos}
\BIBentryALTinterwordspacing
``Ai daos, and three paths to get there | by trent mcconaghy | medium.''
  [Online]. Available:
  \url{https://medium.com/@trentmc0/ai-daos-and-three-paths-to-get-there-cfa0a4cc37b8}
\BIBentrySTDinterwordspacing

\bibitem{ezkl}
\BIBentryALTinterwordspacing
``What is ezkl?'' [Online]. Available: \url{https://docs.ezkl.xyz/}
\BIBentrySTDinterwordspacing

\bibitem{singh2022zero}
N.~Singh, P.~Dayama, and V.~Pandit, ``Zero knowledge proofs towards verifiable
  decentralized ai pipelines,'' in \emph{Financial Cryptography and Data
  Security: 26th International Conference, FC 2022, Grenada, May 2--6, 2022,
  Revised Selected Papers}.\hskip 1em plus 0.5em minus 0.4em\relax Springer,
  2022, pp. 248--275.

\bibitem{openmined_identity}
\BIBentryALTinterwordspacing
``Openmined/pyariesfl: Federated learning on hyperledger aries.'' [Online].
  Available: \url{https://github.com/OpenMined/PyAriesFL}
\BIBentrySTDinterwordspacing

\bibitem{katz1994systems}
M.~L. Katz and C.~Shapiro, ``Systems competition and network effects,''
  \emph{Journal of economic perspectives}, vol.~8, no.~2, pp. 93--115, 1994.

\bibitem{deai_edge}
L.~Cao, ``Decentralized ai: Edge intelligence and smart blockchain, metaverse,
  web3, and desci,'' \emph{IEEE Intelligent Systems}, vol.~37, no.~3, pp.
  6--19, 2022.

\bibitem{unified_framework}
T.~Wang, ``A unified analytical framework for trustable machine learning and
  automation running with blockchain,'' in \emph{2018 IEEE International
  Conference on Big Data (Big Data)}, 2018, pp. 4974--4983.

\bibitem{sabt2015trusted}
M.~Sabt, M.~Achemlal, and A.~Bouabdallah, ``Trusted execution environment: what
  it is, and what it is not,'' in \emph{2015 IEEE Trustcom/BigDataSE/Ispa},
  vol.~1.\hskip 1em plus 0.5em minus 0.4em\relax IEEE, 2015, pp. 57--64.

\bibitem{singla2018machine}
K.~Singla, J.~Bose, and S.~Katariya, ``Machine learning for secure device
  personalization using blockchain,'' in \emph{2018 International Conference on
  Advances in Computing, Communications and Informatics (ICACCI)}.\hskip 1em
  plus 0.5em minus 0.4em\relax IEEE, 2018, pp. 67--73.

\bibitem{dinh2018ai}
T.~N. Dinh and M.~T. Thai, ``Ai and blockchain: A disruptive integration,''
  \emph{Computer}, vol.~51, no.~9, pp. 48--53, 2018.

\bibitem{qiao20206g}
X.~Qiao, Y.~Huang, S.~Dustdar, and J.~Chen, ``6g vision: An ai-driven
  decentralized network and service architecture,'' \emph{IEEE Internet
  Computing}, vol.~24, no.~4, pp. 33--40, 2020.

\bibitem{gupta2020decentralization}
I.~Gupta, ``Decentralization of artificial intelligence: analyzing developments
  in decentralized learning and distributed ai networks,'' \emph{arXiv preprint
  arXiv:1603.04467}, 2020.

\bibitem{zhang2021blockchain}
M.~Zhang, Y.~Li, C.~Zheng, X.~Han, H.~Gu, and H.~Pan, ``Blockchain based global
  financial service platform,'' in \emph{2021 IEEE 19th International
  Conference on Industrial Informatics (INDIN)}.\hskip 1em plus 0.5em minus
  0.4em\relax IEEE, 2021, pp. 1--6.

\bibitem{tian2022blockchain}
R.~Tian, L.~Kong, X.~Min, and Y.~Qu, ``Blockchain for ai: A disruptive
  integration,'' in \emph{2022 IEEE 25th International Conference on Computer
  Supported Cooperative Work in Design (CSCWD)}.\hskip 1em plus 0.5em minus
  0.4em\relax IEEE, 2022, pp. 938--943.

\end{thebibliography}

\section*{Biography}

\begin{IEEEbiography}[{\includegraphics[width=1in,height=1.25in,clip,keepaspectratio]{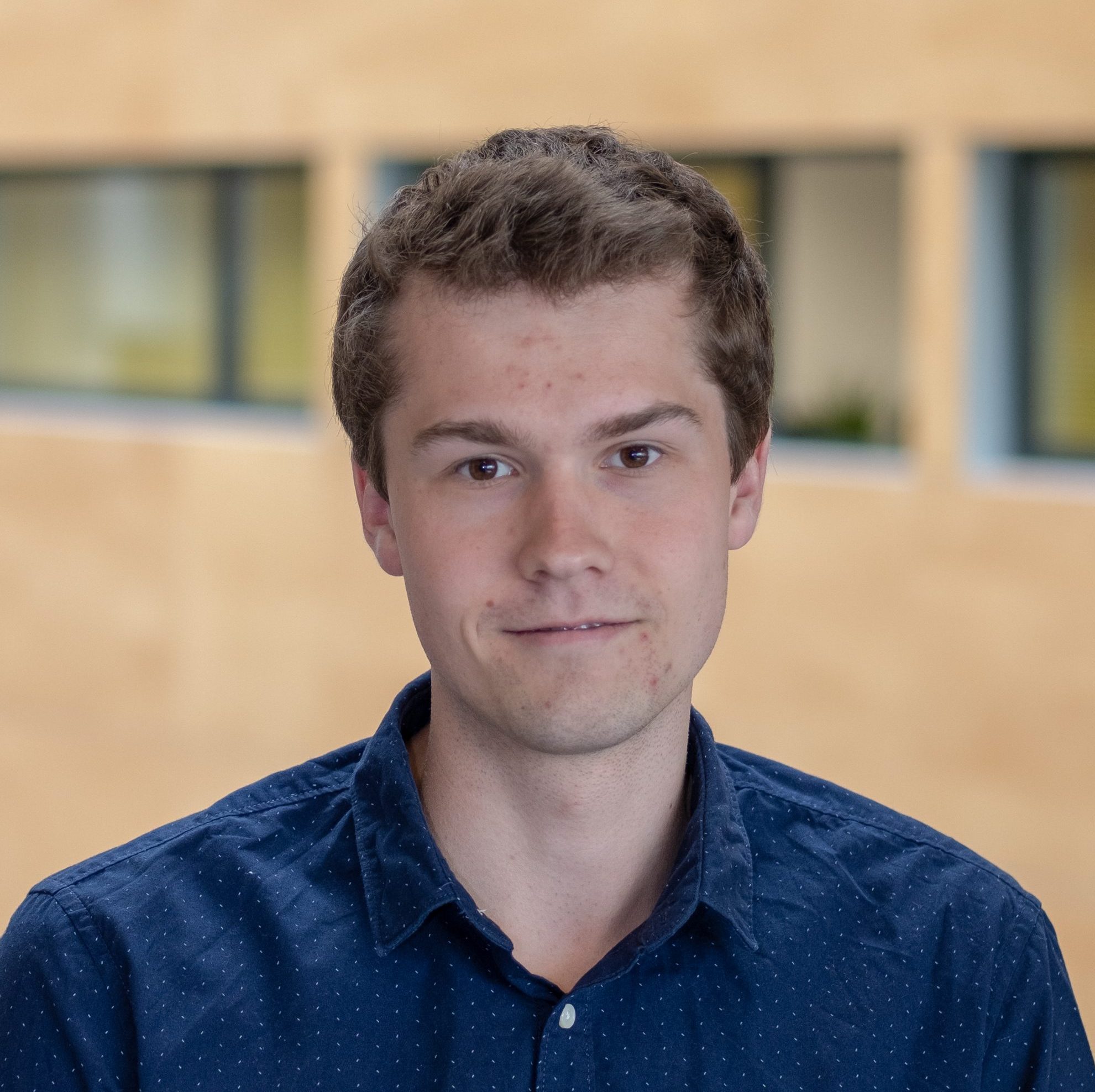}}]{Vid Kersic} is a member of the R\&D group Blockchain Lab:UM, and is working on a Ph.D. with the Faculty of Electrical Engineering and Computer Science, University of Maribor (UM), Maribor, Slovenia. He is currently a Young Researcher with the Institute of Informatics. His current research interests include blockchain, DLTs, decentralized identity, and artificial intelligence. Contact him at vid.kersic@um.si.
\end{IEEEbiography}

\begin{IEEEbiography}[{\includegraphics[width=1in,height=1.25in,clip,keepaspectratio]{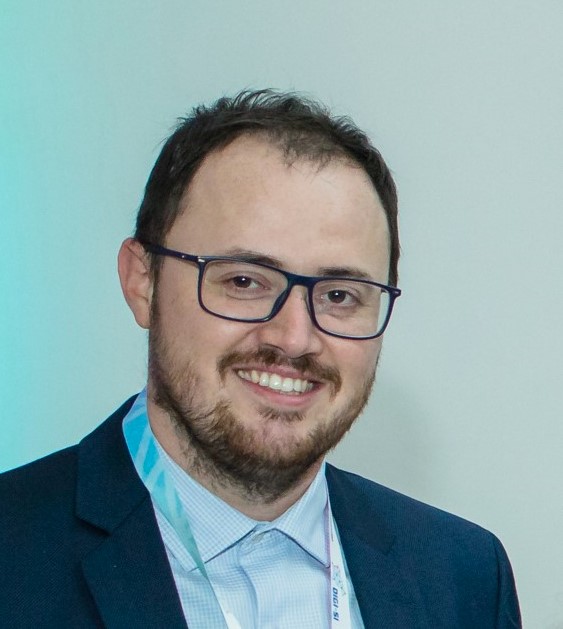}}]{Muhamed Turkanović} was a Managing Director and a CTO of an IT company, from 2013 to 2016. He is currently an Associate Professor at the Faculty of Electrical Engineering and Computer Science, Institute of Informatics, University of Maribor (UM), Slovenia. He has worked there since 2016. He is also the Deputy Head of The Institute of Informatics and Head of Research and Development of the Blockchain Lab:UM as well as the Head of the Slovene EDIH DIGI-SI, UM’s Coordinator of several H2020, HORIZON and/or DIGITAL projects. He has authored several highly cited research papers and, received a patent from EPO, edited several special issues of scientific journals, etc. His current research interests include advanced digital identities, DLTs, database technologies, and applied cryptography.
\end{IEEEbiography}

\clearpage

{
\appendix[Splitted search string]
\label{app:split_ss}

\textbf{ScienceDirect}

\medskip

("p2p ai" \textbf{OR} "p2p artificial intelligence" \textbf{OR} "p2p machine learning" \textbf{OR} "p2p ml" \textbf{OR} "p2p multi-agent system" \textbf{OR} "p2p mas") \textbf{AND} ("network" \textbf{OR} "architecture")

\medskip

("p2p ai" \textbf{OR} "p2p artificial intelligence" \textbf{OR} "p2p machine learning" \textbf{OR} "p2p ml" \textbf{OR} "p2p multi-agent system" \textbf{OR} "p2p mas") \textbf{AND} ("platform" \textbf{OR} "services" \textbf{OR} "marketplace")

\medskip

("peer-to-peer ai" \textbf{OR} "peer-to-peer artificial intelligence" \textbf{OR} "peer-to-peer machine learning" \textbf{OR} "peer-to-peer ml" \textbf{OR} "peer-to-peer multi-agent system" \textbf{OR} "peer-to-peer mas") \textbf{AND} ("network" \textbf{OR} "architecture")

\medskip

("peer-to-peer ai" \textbf{OR} "peer-to-peer artificial intelligence" \textbf{OR} "peer-to-peer machine learning" \textbf{OR} "peer-to-peer ml" \textbf{OR} "peer-to-peer multi-agent system" \textbf{OR} "peer-to-peer mas") \textbf{AND} ("platform" \textbf{OR} "services" \textbf{OR} "marketplace")

\medskip

("decentralized ai" \textbf{OR} "decentralized artificial intelligence" \textbf{OR} "decentralized machine learning" \textbf{OR} "decentralized ml" \textbf{OR} "decentralized multi-agent system" \textbf{OR} "decentralized mas") \textbf{AND} ("network" \textbf{OR} "architecture")

\medskip

("decentralized ai" \textbf{OR} "decentralized artificial intelligence" \textbf{OR} "decentralized machine learning" \textbf{OR} "decentralized ml" \textbf{OR} "decentralized multi-agent system" \textbf{OR} "decentralized mas") \textbf{AND} ("platform" \textbf{OR} "services" \textbf{OR} "marketplace")

\medskip

Constraints:
\begin{itemize}
    \item Title, abstract or author-specified keywords
    \item Subject areas: computer science and engineering
    \item Publication title: information sciences, expert systems with applications, and technological forecasting and social change
\end{itemize}

\medskip

\textbf{SpringerLink}

\medskip

Search string were the same as for ScienceDirect.

\medskip

Constraints:
\begin{itemize}
    \item Do not include preview-only content
\end{itemize}

\medskip

\textbf{IEEE Xplore Digital Library}

\medskip

(("Abstract":"p2p ai" \textbf{OR} "Abstract":"p2p artificial intelligence" \textbf{OR} "Abstract":"p2p machine learning" \textbf{OR} "Abstract":"p2p ml" \textbf{OR} "Abstract":"p2p multi-agent system" \textbf{OR} "Abstract":"p2p mas" \textbf{OR} "Abstract":"peer-to-peer ai" \textbf{OR} "Abstract":"peer-to-peer artificial intelligence" \textbf{OR} "Abstract":"peer-to-peer machine learning" \textbf{OR} "Abstract":"peer-to-peer ml" \textbf{OR} "Abstract":"peer-to-peer multi-agent system" \textbf{OR} "Abstract":"peer-to-peer mas" \textbf{OR} "Abstract":"decentralized ai" \textbf{OR} "Abstract":"decentralized artificial intelligence" \textbf{OR} "Abstract":"decentralized machine learning" \textbf{OR} "Abstract":"decentralized ml" \textbf{OR} "Abstract":"decentralized multi-agent system" \textbf{OR} "Abstract":"decentralized mas") \textbf{AND} ("Abstract":"network" \textbf{OR} "Abstract":"architecture")) \textbf{OR} (("Document Title":"p2p ai" \textbf{OR} "Document Title":"p2p artificial intelligence" \textbf{OR} "Document Title":"p2p machine learning" \textbf{OR} "Document Title":"p2p ml" \textbf{OR} "Document Title":"p2p multi-agent system" \textbf{OR} "Document Title":"p2p mas" \textbf{OR} "Document Title":"peer-to-peer ai" \textbf{OR} "Document Title":"peer-to-peer artificial intelligence" \textbf{OR} "Document Title":"peer-to-peer machine learning" \textbf{OR} "Document Title":"peer-to-peer ml" \textbf{OR} "Document Title":"peer-to-peer multi-agent system" \textbf{OR} "Document Title":"peer-to-peer mas" \textbf{OR} "Document Title":"decentralized ai" \textbf{OR} "Document Title":"decentralized artificial intelligence" \textbf{OR} "Document Title":"decentralized machine learning" \textbf{OR} "Document Title":"decentralized ml" \textbf{OR} "Document Title":"decentralized multi-agent system" \textbf{OR} "Document Title":"decentralized mas") \textbf{AND} ("Document Title":"network" \textbf{OR} "Document Title":"architecture")) \textbf{OR} (("Publication Title":"p2p ai" \textbf{OR} "Publication Title":"p2p artificial intelligence" \textbf{OR} "Publication Title":"p2p machine learning" \textbf{OR} "Publication Title":"p2p ml" \textbf{OR} "Publication Title":"p2p multi-agent system" \textbf{OR} "Publication Title":"p2p mas" \textbf{OR} "Publication Title":"peer-to-peer ai" \textbf{OR} "Publication Title":"peer-to-peer artificial intelligence" \textbf{OR} "Publication Title":"peer-to-peer machine learning" \textbf{OR} "Publication Title":"peer-to-peer ml" \textbf{OR} "Publication Title":"peer-to-peer multi-agent system" \textbf{OR} "Publication Title":"peer-to-peer mas" \textbf{OR} "Publication Title":"decentralized ai" \textbf{OR} "Publication Title":"decentralized artificial intelligence" \textbf{OR} "Publication Title":"decentralized machine learning" \textbf{OR} "Publication Title":"decentralized ml" \textbf{OR} "Publication Title":"decentralized multi-agent system" \textbf{OR} "Publication Title":"decentralized mas") \textbf{AND} ("Publication Title":"network" \textbf{OR} "Publication Title":"architecture")) \textbf{OR} (("Index Terms":"p2p ai" \textbf{OR} "Index Terms":"p2p artificial intelligence" \textbf{OR} "Index Terms":"p2p machine learning" \textbf{OR} "Index Terms":"p2p ml" \textbf{OR} "Index Terms":"p2p multi-agent system" \textbf{OR} "Index Terms":"p2p mas" \textbf{OR} "Index Terms":"peer-to-peer ai" \textbf{OR} "Index Terms":"peer-to-peer artificial intelligence" \textbf{OR} "Index Terms":"peer-to-peer machine learning" \textbf{OR} "Index Terms":"peer-to-peer ml" \textbf{OR} "Index Terms":"peer-to-peer multi-agent system" \textbf{OR} "Index Terms":"peer-to-peer mas" \textbf{OR} "Index Terms":"decentralized ai" \textbf{OR} "Index Terms":"decentralized artificial intelligence" \textbf{OR} "Index Terms":"decentralized machine learning" \textbf{OR} "Index Terms":"decentralized ml" \textbf{OR} "Index Terms":"decentralized multi-agent system" \textbf{OR} "Index Terms":"decentralized mas") \textbf{AND} ("Index Terms":"network" \textbf{OR} "Index Terms":"architecture"))

\medskip

(("Abstract":"p2p ai" \textbf{OR} "Abstract":"p2p artificial intelligence" \textbf{OR} "Abstract":"p2p machine learning" \textbf{OR} "Abstract":"p2p ml" \textbf{OR} "Abstract":"p2p multi-agent system" \textbf{OR} "Abstract":"p2p mas" \textbf{OR} "Abstract":"peer-to-peer ai" \textbf{OR} "Abstract":"peer-to-peer artificial intelligence" \textbf{OR} "Abstract":"peer-to-peer machine learning" \textbf{OR} "Abstract":"peer-to-peer ml" \textbf{OR} "Abstract":"peer-to-peer multi-agent system" \textbf{OR} "Abstract":"peer-to-peer mas" \textbf{OR} "Abstract":"decentralized ai" \textbf{OR} "Abstract":"decentralized artificial intelligence" \textbf{OR} "Abstract":"decentralized machine learning" \textbf{OR} "Abstract":"decentralized ml" \textbf{OR} "Abstract":"decentralized multi-agent system" \textbf{OR} "Abstract":"decentralized mas") \textbf{AND} ("Abstract":"platform" \textbf{OR} "Abstract":"services" \textbf{OR} "Abstract":"marketplace")) \textbf{OR} (("Document Title":"p2p ai" \textbf{OR} "Document Title":"p2p artificial intelligence" \textbf{OR} "Document Title":"p2p machine learning" \textbf{OR} "Document Title":"p2p ml" \textbf{OR} "Document Title":"p2p multi-agent system" \textbf{OR} "Document Title":"p2p mas" \textbf{OR} "Document Title":"peer-to-peer ai" \textbf{OR} "Document Title":"peer-to-peer artificial intelligence" \textbf{OR} "Document Title":"peer-to-peer machine learning" \textbf{OR} "Document Title":"peer-to-peer ml" \textbf{OR} "Document Title":"peer-to-peer multi-agent system" \textbf{OR} "Document Title":"peer-to-peer mas" \textbf{OR} "Document Title":"decentralized ai" \textbf{OR} "Document Title":"decentralized artificial intelligence" \textbf{OR} "Document Title":"decentralized machine learning" \textbf{OR} "Document Title":"decentralized ml" \textbf{OR} "Document Title":"decentralized multi-agent system" \textbf{OR} "Document Title":"decentralized mas") \textbf{AND} ("Document Title":"platform" \textbf{OR} "Document Title":"services" \textbf{OR} "Document Title":"marketplace")) \textbf{OR} (("Publication Title":"p2p ai" \textbf{OR} "Publication Title":"p2p artificial intelligence" \textbf{OR} "Publication Title":"p2p machine learning" \textbf{OR} "Publication Title":"p2p ml" \textbf{OR} "Publication Title":"p2p multi-agent system" \textbf{OR} "Publication Title":"p2p mas" \textbf{OR} "Publication Title":"peer-to-peer ai" \textbf{OR} "Publication Title":"peer-to-peer artificial intelligence" \textbf{OR} "Publication Title":"peer-to-peer machine learning" \textbf{OR} "Publication Title":"peer-to-peer ml" \textbf{OR} "Publication Title":"peer-to-peer multi-agent system" \textbf{OR} "Publication Title":"peer-to-peer mas" \textbf{OR} "Publication Title":"decentralized ai" \textbf{OR} "Publication Title":"decentralized artificial intelligence" \textbf{OR} "Publication Title":"decentralized machine learning" \textbf{OR} "Publication Title":"decentralized ml" \textbf{OR} "Publication Title":"decentralized multi-agent system" \textbf{OR} "Publication Title":"decentralized mas") \textbf{AND} ("Publication Title":"platform" \textbf{OR} "Publication Title":"services" \textbf{OR} "Publication Title":"marketplace")) \textbf{OR} (("Index Terms":"p2p ai" \textbf{OR} "Index Terms":"p2p artificial intelligence" \textbf{OR} "Index Terms":"p2p machine learning" \textbf{OR} "Index Terms":"p2p ml" \textbf{OR} "Index Terms":"p2p multi-agent system" \textbf{OR} "Index Terms":"p2p mas" \textbf{OR} "Index Terms":"peer-to-peer ai" \textbf{OR} "Index Terms":"peer-to-peer artificial intelligence" \textbf{OR} "Index Terms":"peer-to-peer machine learning" \textbf{OR} "Index Terms":"peer-to-peer ml" \textbf{OR} "Index Terms":"peer-to-peer multi-agent system" \textbf{OR} "Index Terms":"peer-to-peer mas" \textbf{OR} "Index Terms":"decentralized ai" \textbf{OR} "Index Terms":"decentralized artificial intelligence" \textbf{OR} "Index Terms":"decentralized machine learning" \textbf{OR} "Index Terms":"decentralized ml" \textbf{OR} "Index Terms":"decentralized multi-agent system" \textbf{OR} "Index Terms":"decentralized mas") \textbf{AND} ("Index Terms":"platform" \textbf{OR} "Index Terms":"services" \textbf{OR} "Index Terms":"marketplace"))

\medskip

Constraints:
\begin{itemize}
    \item Abstract, Index Terms, Publication Title, and Document Title
\end{itemize}

\medskip

\textbf{ACM Digital Library}

\medskip

Search strings were the same as for ScienceDirect.

\medskip

Constraints:
\begin{itemize}
    \item Anywhere
    \item Open Access
\end{itemize}

\medskip

\textbf{Web of Science}

\medskip

Search strings were the same as for ScienceDirect.

\medskip

Constraints:
\begin{itemize}
    \item All Fields
\end{itemize}

\medskip

\textbf{Google Scholar}

\medskip

Search strings were the same as for ScienceDirect. We also ran more loose search strings due to being limited to only the first ten pages, e.g., only "decentralized artificial intelligence," which returned positive results and articles that were included in the end.

}

\clearpage
\onecolumn

{
\appendix[All identified articles]
\label{app:all_articles}

\begin{longtable}{|c|c|ccccccccccccc|}
\caption{IDENTIFIED BUILDING BLOCKS FOR ALL WORKS. A - registry, B - incentivization, C - marketplace, D - reputation, E - ontology, F - discoverability, G - training (computation), H - inference (computation), I - ownership, J - data, K - governance, L - cryptography and privacy, M - identity.} \\
\hline
\multirow{2}{*}{\textbf{Article}}         & \multirow{2}{*}{\textbf{Database}} & \multicolumn{13}{c|}{\textbf{Building Blocks}}                                                                                                                                                                                                                                                                                                                                                                                     \\ \cline{3-15} 
                                          &                                    & \multicolumn{1}{c|}{\textbf{A}} & \multicolumn{1}{c|}{\textbf{B}} & \multicolumn{1}{c|}{\textbf{C}} & \multicolumn{1}{c|}{\textbf{D}} & \multicolumn{1}{c|}{\textbf{E}} & \multicolumn{1}{c|}{\textbf{F}} & \multicolumn{1}{c|}{\textbf{G}} & \multicolumn{1}{c|}{\textbf{H}} & \multicolumn{1}{c|}{\textbf{I}} & \multicolumn{1}{c|}{\textbf{J}} & \multicolumn{1}{c|}{\textbf{K}} & \multicolumn{1}{c|}{\textbf{L}} & \textbf{M} \\ \hline
\endfirsthead
\endhead
\cite{miiller1990decentralized}           & Google Scholar                     & \multicolumn{1}{c|}{ }          & \multicolumn{1}{c|}{ }          & \multicolumn{1}{c|}{ }          & \multicolumn{1}{c|}{ }          & \multicolumn{1}{c|}{x}          & \multicolumn{1}{c|}{ }          & \multicolumn{1}{c|}{ }          & \multicolumn{1}{c|}{ }          & \multicolumn{1}{c|}{ }          & \multicolumn{1}{c|}{ }          & \multicolumn{1}{c|}{ }          & \multicolumn{1}{c|}{ }          &            \\ \hline
\cite{tran2008preliminary}                & SpringerLink                       & \multicolumn{1}{c|}{ }          & \multicolumn{1}{c|}{ }          & \multicolumn{1}{c|}{ }          & \multicolumn{1}{c|}{ }          & \multicolumn{1}{c|}{x}          & \multicolumn{1}{c|}{ }          & \multicolumn{1}{c|}{ }          & \multicolumn{1}{c|}{ }          & \multicolumn{1}{c|}{ }          & \multicolumn{1}{c|}{ }          & \multicolumn{1}{c|}{ }          & \multicolumn{1}{c|}{ }          &            \\ \hline
\cite{gorodetskii2008development}         & SpringerLink                       & \multicolumn{1}{c|}{x}          & \multicolumn{1}{c|}{ }          & \multicolumn{1}{c|}{ }          & \multicolumn{1}{c|}{ }          & \multicolumn{1}{c|}{x}          & \multicolumn{1}{c|}{x}          & \multicolumn{1}{c|}{ }          & \multicolumn{1}{c|}{ }          & \multicolumn{1}{c|}{ }          & \multicolumn{1}{c|}{ }          & \multicolumn{1}{c|}{ }          & \multicolumn{1}{c|}{ }          &            \\
\hline
\end{longtable}

\clearpage
\onecolumn

\begin{longtable}{|c|c|ccc|c|}
\caption{IDENTIFIED FEATURES AND CHALLENGES FOR ALL WORKS.} \\
\hline
\multirow{2}{*}{\textbf{Article}}                              & \multirow{2}{*}{\textbf{Database}} & \multicolumn{3}{c|}{\textbf{Features}}                                                                                                                                & \textbf{Challenges}                                                       \\ \cline{3-6} 
                                                               &                                    & \multicolumn{1}{c|}{\textbf{\begin{tabular}[c]{@{}c@{}}Standards and\\ Specifications\end{tabular}}} & \multicolumn{1}{c|}{\textbf{Security}} & \textbf{Transparency} & \textbf{\begin{tabular}[c]{@{}c@{}}Updating Global\\ Models\end{tabular}} \\ \hline
\endfirsthead
\endhead
\cite{miiller1990decentralized}               & Google Scholar                     & \multicolumn{1}{c|}{x}                                                                               & \multicolumn{1}{c|}{ }                 &                       &                                                                           \\ \hline
\cite{tran2008preliminary}                    & SpringerLink                       & \multicolumn{1}{c|}{x}                                                                               & \multicolumn{1}{c|}{ }                 &                       &                                                                           \\ \hline
\cite{gorodetskii2008development}             & SpringerLink                       & \multicolumn{1}{c|}{x}                                                                               & \multicolumn{1}{c|}{x}                 &                       &                                                                           \\ \hline
\cite{numerai}                                & Gray literature                    & \multicolumn{1}{c|}{ }                                                                               & \multicolumn{1}{c|}{ }                 & x                     & x                                                                         \\ \hline
\cite{openmined_identity}                    & Gray literature                    & \multicolumn{1}{c|}{ }                                                                               & \multicolumn{1}{c|}{ }                 & x                     & x                                                                         \\
\hline
\end{longtable}

\clearpage
\twocolumn

}

\end{document}